\documentclass{article}
\usepackage{arxiv}

\usepackage[utf8]{inputenc} 
\usepackage[T1]{fontenc}    

\usepackage{booktabs}       
\usepackage{amsmath}
\usepackage{amsfonts}       
\usepackage{nicefrac}       
\usepackage{microtype}      
\usepackage{lipsum}		
\usepackage{graphicx}
\usepackage{doi}

\usepackage{comment}
\usepackage{algorithm}
\usepackage{algpseudocode}
\usepackage{xcolor}
\usepackage{setspace}
\usepackage{cite}
\usepackage{lscape}
\usepackage{indentfirst}
\usepackage{latexsym}
\usepackage{multirow}
\usepackage{tabls}
\usepackage{wrapfig}
\usepackage{longtable}
\usepackage{supertabular}
\usepackage{caption}
\usepackage{subcaption}
\usepackage{authblk}
\usepackage{url}            
\usepackage{hyperref}       

\title{Incorporating Background Knowledge in Symbolic Regression using a Computer Algebra System}

\author[2]{\normalsize Charles Fox}
\author[1]{\normalsize Neil Tran}
\author[1]{\normalsize Nikki Nacion}
\author[1]{\normalsize Samiha Sharlin}
\author[1,2]{\normalsize  Tyler R. Josephson}

\affil[1]{\normalsize Department of Chemical, Biochemical, and Environmental Engineering, University of Maryland Baltimore County, \authorcr
1000 Hilltop Circle, Baltimore, MD 21250
}
\affil[2]{\normalsize Department of Computer Science and Electrical Engineering, University of Maryland Baltimore County, \authorcr
1000 Hilltop Circle, Baltimore, MD 21250}

\date{}

\renewcommand{\shorttitle}{Background Knowledge in Symbolic Regression}
\hypersetup{
pdftitle={Incorporating Background Knowledge in Symbolic Regression using a Computer Algebra System},
pdfauthor={Charles Fox},
}

\begin{document}
\maketitle

\begin{abstract}
Symbolic Regression (SR) can generate interpretable, concise expressions that fit a given dataset, allowing for more human understanding of the structure than black-box approaches. The addition of background knowledge (in the form of symbolic mathematical constraints) allows for the generation of expressions that are meaningful with respect to theory while also being consistent with data. We specifically examine the addition of constraints to traditional genetic algorithm (GA) based SR (PySR) as well as a Markov-chain Monte Carlo (MCMC) based Bayesian SR architecture (Bayesian Machine Scientist), and apply these to rediscovering adsorption equations from experimental, historical datasets.
We find that, while hard constraints prevent GA and MCMC SR from searching, soft constraints can lead to improved performance both in terms of search effectiveness and model meaningfulness, with computational costs increasing by about an order-of-magnitude. If the constraints do not correlate well with the dataset or expected models, they can hinder the search of expressions. We find incorporating these constraints in Bayesian SR (as the Bayesian prior) is better than by modifying the fitness function in the GA. 
\end{abstract}

\section{Introduction}
\subsection{Symbolic Regression for Scientific Discovery}

Symbolic Regression (SR) generates mathematical expressions that are optimized for complexity and accuracy to a given dataset. 
Since John Koza pioneered the paradigm of programming by means of natural selection, 
many applications for SR in scientific discovery have emerged \cite{koza_genetic_1992}. 
Unlike other applications of machine learning techniques, scientific research demands explanation and verification, both of which are made more feasible by the generation of human-interpretable mathematical models (as opposed to fitting a model with thousands of parameters) \cite{oviedo2022interpretable, zhong2022explainable, esterhuizen2022interpretable}.  Furthermore, SR can be effective even with very small datasets ($\sim$10 items) such as those produced by difficult or expensive experiments which are not easily repeated. The mathematical expressions produced by SR can easily be extrapolated to untested or otherwise unreachable domains within a dataset (such as extreme pressures or temperatures). 

For decades, SR has discovered interesting models from data in many applications including inferring process models at the Dow Chemical Company \cite{kordon_application_2006}, rainfall-runoff modeling \cite{savic_genetic_1999}, and rediscovering equations for double-pendulum motion \cite{schmidt_distilling_2009}. Symbolic regression has been applied across all scales of scientific investigation, including the atomistic (interatomic potentials \cite{hernandez_fast_2019}), macroscopic (computational fluid dynamics \cite{ansari_iterative_2022}), and cosmological (dark matter overdensity \cite{cranmer_discovering_2020}) scales. Some techniques facilitate search through billions of candidate expressions, such as the space of nonlinear descriptors of material properties \cite{ouyang_sisso_2018}. 
While most applications of SR in science focus on identifying empirical patterns in data, such "data-only" approaches do not account for potential insights from background theory. In fact, some SR works emphasize their capabilities of discovery ``without any prior knowledge about physics, kinematics, or geometry'' \cite{schmidt_distilling_2009}. Nonetheless, we posit that prior knowledge need not be discarded, and in this work, we explore how theory may be incorporated into symbolic regression to demonstrate machine learning in the context of background knowledge.

\subsection{Incorporating Background Knowledge into Symbolic Regression}



One particularly important step towards effective use of SR in specific domains is the addition of prior knowledge. This step has the potential to take a general purpose SR algorithm and use it to find novel models with physical meaning. For example, AI-DARWIN is uses prior knowledge of chemical reaction mechanisms in the form of predefined functions that a genetic algorithm may use in its search of equation space, ensuring that each generated model is mechanistically meaningful \cite{chakraborty_ai-darwin_2021}. This approach specifically encodes the prior knowledge in the form of functions available instead of limitations on functions generated. 
In another recent example, Engle and Sahinidis use a deterministic symbolic regression algorithm that constrains the space of possible equations, not to those constructed from a library of meaningful function components, but to those functions that obey derivative constraints from theory. This improves the quality of generated expressions for thermodynamic equations of state \cite{engle_deterministic_2022}.
Another approach to incorporating background knowledge in symbolic regression is the Bayesian Machine Scientist (BMS) \cite{guimera_bayesian_2020}. BMS rigorously incorporates background knowledge in the form of a Bayesian prior on symbolic expressions; expressions are \emph{a priori} more likely if their distribution of mathematical operators aligns with the distribution of operators in a corpus of prominent equations. However, their approach to the Bayesian prior does not incorporate meaning from particular scientific domains.

Checking consistency of equations \emph{after} the search is complete is also possible. Previously, we showed that generated expressions can be compared to rich background knowledge (expressed as axioms for the environment under study), by posing generated expressions as conjectures to an automated theorem prover (ATP) \cite{cornelio_ai_2021}. However, state-of-the-art ATPs are too slow to incorporate this logical check as symbolic expressions are generated, and therefore cannot be easily used to bias the search for equations in light of that background knowledge. Moreover, translating scientific theories into a computer-interpretable form is not straightforward.

We address these specific drawbacks by combining symbolic regression systems (both genetic algorithm and Bayesian approaches) with a computer algebra system (CAS) that checks constraints as an equation search is conducted.
This is similar to Logic Guided Genetic Algorithms (LGGA), which uses ``auxiliary truths'' (ATs) corresponding to datasets in order to weigh items in a dataset as well as augment it with more information \cite{ashok_logic_2021}.  LGGA follows an iterative approach of training an arbitrary genetic algorithm with some dataset, augmenting that dataset with ATs, and training that algorithm again with more informative data. An important distinction between our work and LGGA is that the dataset is not altered in any way and the addition of extra information is performed during the execution of the GA. Another related approach is shape-constrained symbolic regression \cite{kronbergerShapeconstrainedSymbolicRegression2022, haiderShapeconstrainedMultiobjectiveGenetic2023}, in which constraints on function shape (e.g. partial derivatives and monotonicity) are incorporated into symbolic regression using an efficient application of integer arithmetic. Our approach considers a broader range of constraints (any that can be defined and checked by the CAS), and considers both a genetic programming and Bayesian symbolic regression approach.

\subsection{Adsorption}

Adsorption, the phenomenon in which molecules bind to a surface, enables chemical processes including carbon capture, humidity control, removal of harmful pollutants from water, and hydrogen production \cite{ben-mansour_carbon_2016} \cite{ritter_stateart_2007} \cite{stenzel_remove_1993} \cite{ruthven_principles_1984}. Models of adsorption enable prediction and design of engineered adsorption processes, and many have been proposed over the years (selected equations are shown in Table~\ref{tab:adsorptionIsotherms}) \cite{limousin2007sorption, foo2010insights, ayawei2017modelling, wang_adsorption_2020}. These models relate the amount adsorbed at equilibrium as a function of pressure or concentration and are commonly expressed as equations that are either empirical or derived from theory. For example, the Freundlich isotherm \cite{freundlich1906adsorption}, is an empirical function designed to fit observed data, the Langmuir \cite{langmuir_adsorption_1918} and BET \cite{brunauer_adsorption_1938} isotherms are derived from physical models, and the Sips \cite{sips1948structure} isotherm is Langmuir-inspired with empirical terms added for fitting flexibility. 
We wonder, ``What kinds of models could be generated by a machine learning system, and what role can background knowledge play in the search for accurate and meaningful expressions?''

\begin{table}[ht]
\centering
\caption{Some well-known isotherms written as SR might find them, and their complexities.}
\label{tab:adsorptionIsotherms}
\begin{tabular}{|l|c|c|c|} 
\hline
\textbf{Isotherm}  & \textbf{Literature Expression}                                                                & \textbf{Symbolic Regression Form}       & \textbf{SR Complexity}  \\ 
\hline
Langmuir \cite{langmuir_adsorption_1918}          & $\frac{q_{max}K_{eq}p}{1+K_{eq}p}$                                                            & $\frac{c_1p}{c_2+p}$                    & 7                       \\ 
\hline
Dual-Site Langmuir \cite{langmuir_adsorption_1918} & $\frac{q^{a}_{max}K^{a}_{eq}p}{1+K^{a}_{eq}p} + \frac{q^{b}_{max}K^{b}_{eq}p}{1+K^{b}_{eq}p}$ & $\frac{c_1p}{c_2+p}+\frac{c_3p}{c_4+p}$ & 15                      \\ 
\hline
BET \cite{brunauer_adsorption_1938}                & $\frac{v_m*c*(p/p_0)}{(1-p/p_0)*(1+ (c-1)p/p_0)}$                                         & $\frac{c_1p}{p^2 + c2p + c3}$  & 13                      \\ 
\hline
Freundlich \cite{freundlich1906adsorption}        & $c_1p^\frac{1}{n}$                                                                            & $c_1p^{c_2}$                            & 5                       \\ 
\hline
Sips \cite{sips1948structure}              & $\frac{c_1p^{\frac{1}{n}}}{1+c_1p^{\frac{1}{n}}}$                                             & $\frac{p^{c_2}}{c_1+p^{c_2}}$       & 9                      \\
\hline
\end{tabular}
\end{table}

\newpage
\subsection{Thermodynamic Constraints}  \label{sec:OriginalThermoConstraints}
We consider models to be more \emph{meaningful} when they satisfy thermodynamic constraints on the functional forms appropriate for modeling these phenomena. That is, a random equation that fits data, but does not approach zero loading correctly, is less trustworthy outside the training data than an equation constrained to follow thermodynamics. We have identified three constraints relevant for single-component adsorption \cite{cornelio_ai_2021}:

\begin{gather}
    \lim_{p\to0} f(p)=0 \\
    \lim_{p\to0} f'(p)<\infty \\
    \forall p >0 \qquad f'(p)\geq0
\end{gather}


Constraint 1 ensures that, in the limit of zero pressure, all molecules must desorb, and loading cannot be negative. Constraint 2 requires that in the limit of zero pressure, the slope of the isotherm must be a positive finite constant. Talu and Myers show that, as pressure approaches zero, the slope of the adsorption isotherm equals the adsorption second virial coefficient $B_{1S}$, which characterizes the interaction between one molecule and the surface, and must be a finite positive number \cite{talu_rigorous_1988} \cite{toth_consequences_1997}:  
\begin{gather}
    \lim_{p\to0} \frac{df}{dp} = \frac{B_{1S}}{RT} = c
\end{gather} 
Constraint 3 requires that loading does not decrease with increasing pressure (the isotherm is monotonically non-decreasing) for all ($\forall$) positive values of pressure. Note that this does not hold for mixture adsorption (in which competition plays a role), nor in BET adsorption, which exhibits a discontinuity at the saturation pressure, instead of a monotonic increase.


\subsection{PySR: Symbolic Regression using Genetic Algorithms}  
PySR, Python for Symbolic Regression, is a Python library that uses a genetic algorithm for symbolic regression \cite{cranmer_milescranmerpysr_2021}. 
PySR is a Python wrapper that calls a Julia library by the same author, SymbolicRegression.jl (SR.jl), for numerical performance.
Due to the nature of the modifications needed to the algorithm for this work, the base Julia library was used, but all added functionality should be inherited by the Python wrapper library as well.

The basic premise is that one or more populations of models move towards more optimal solutions via random mutations.  At each generation, some members of a population are removed based on their fitness, age, or some other criteria (PySR replaces the oldest members). Beneficial solutions are encouraged by having more optimal members of a population mutate and reproduce.

\begin{figure}[H]
    \centering
    \includegraphics[width=\textwidth]{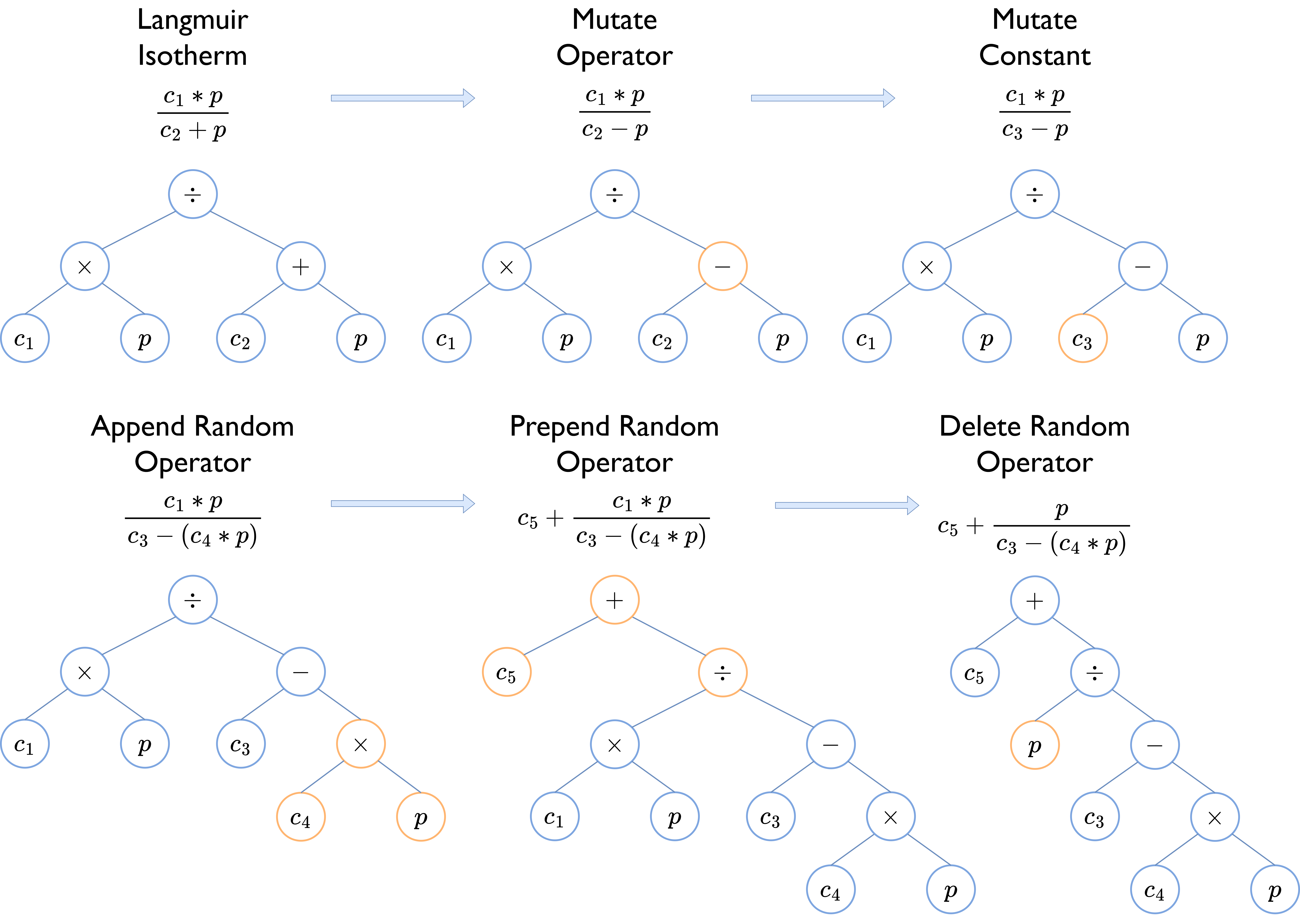}
    \caption{All mutations (except for random tree generation and simplification) in PySR in succession (read from left to right, top to bottom).  Changes from each previous expression tree are shown in orange.}
    \label{fig:PySRMutations}
\end{figure}

\begin{figure}[H]
    \centering
    \includegraphics[trim={0 0 0 2.1cm}, clip, width=0.5\textwidth]{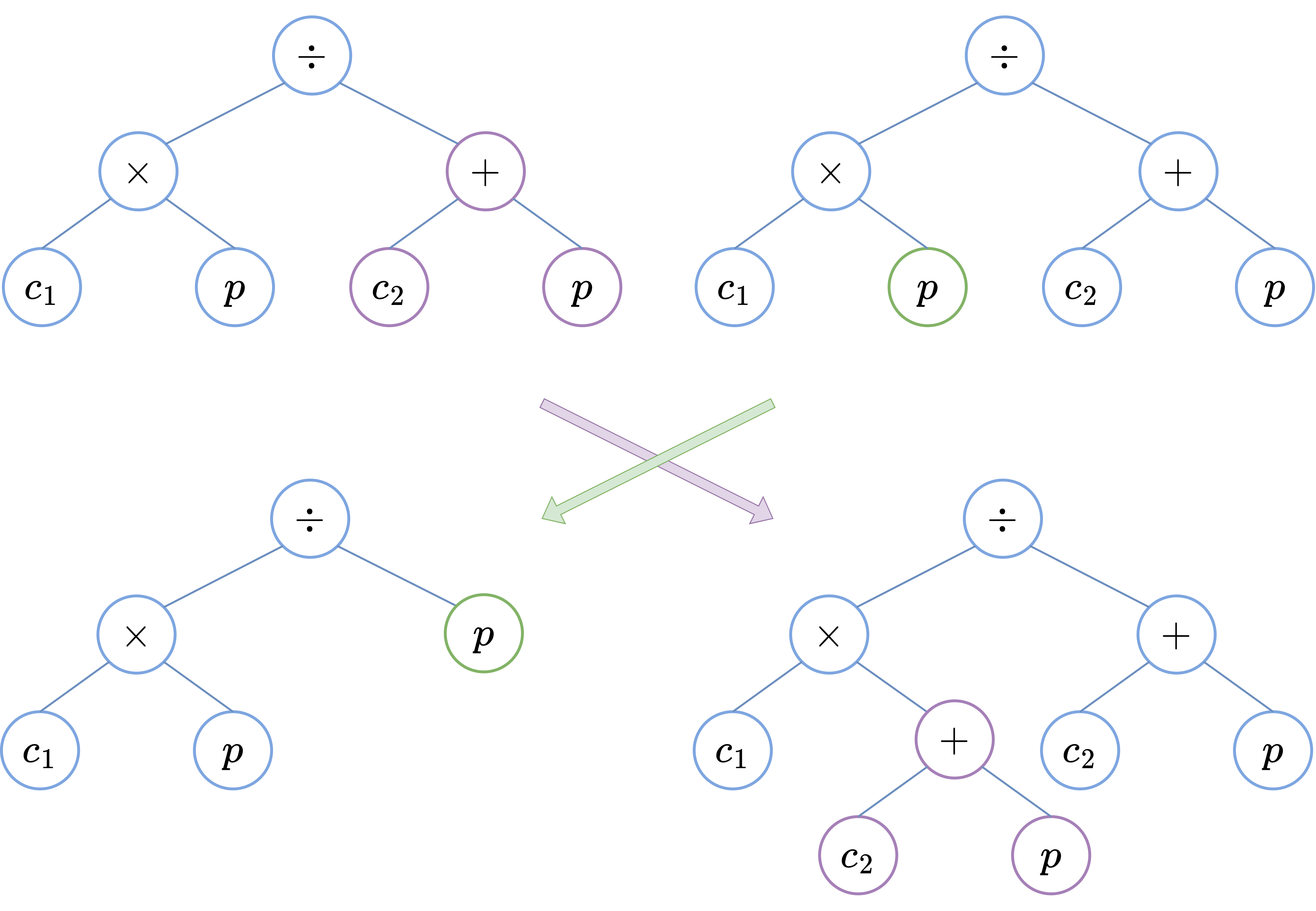}
    \caption{An example of the crossover mutation between two expression trees.}
    \label{fig:PySRCrossover}
\end{figure}

Changes include mutating a single constant or operator, simplifying the expression, or performing crossover between two expressions (Fig. \ref{fig:PySRMutations} and Fig. \ref{fig:PySRCrossover}). 
PySR uses multiple populations in a method similar to the island methodology \cite{konfrst_parallel_2004}.
This aims to allow for specialization by separately evolving unique populations, occasionally allowing some members to move between them to share that specialization.
Specifically, PySR implements the so-called Hall of Fame (HOF), which is a Pareto front built from the best members across each population.
After a number of generations, each population submits its top 10 best members (based on score) which are then compared and pared down via Pareto front.  Expressions that remain in the HOF are used for future mutations in each of the populations.




\subsection{Bayesian Symbolic Regression} 
\begin{figure}[H]
    \centering
    \includegraphics[width=\textwidth]{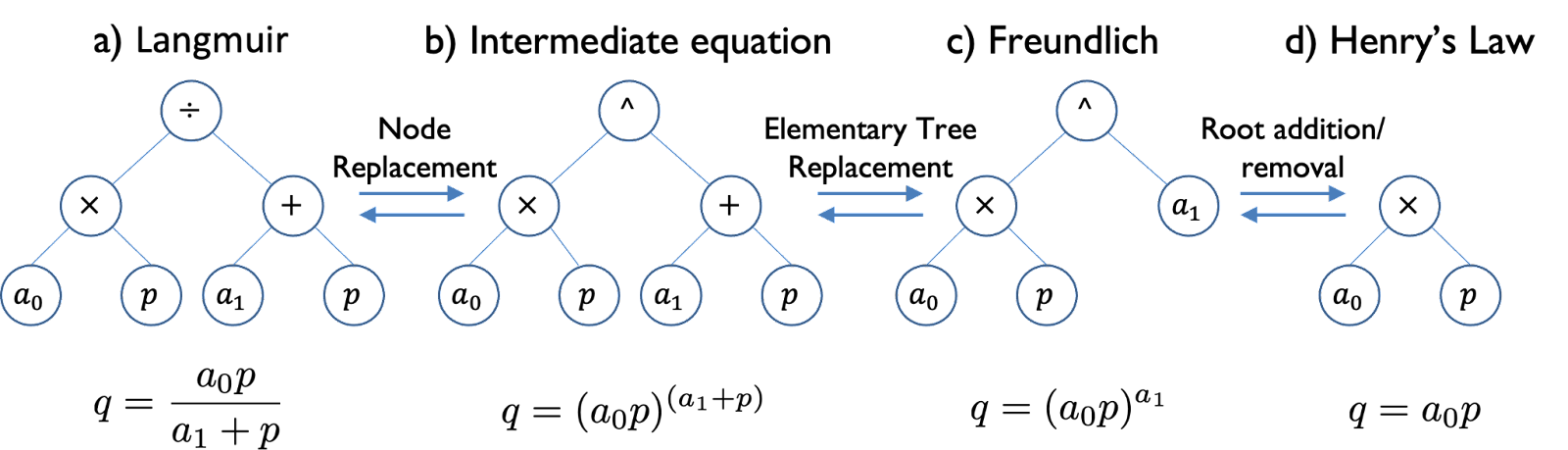}
    \caption{Illustrating the moves available to the BMS algorithm, as applied to adsorption equations. In contrast to the mutations available in PySR, these transformations satisfy detailed balance \cite{guimera_bayesian_2020}.}
    \label{fig:BMSMoves}
\end{figure}

The Bayesian Machine Scientist (BMS) by Guimera et. al. \cite{guimera_bayesian_2020} approaches symbolic regression from a Bayesian perspective. 
Bayesian Symbolic Regression (BSR) frames the search for accurate, concise and informed models as sampling the marginal posterior distribution of symbolic models with respect to a prior and fit to a dataset. Markov chain Monte Carlo (MC) is used to generate new expression trees (Fig.~\ref{fig:BMSMoves}), which are accepted or rejected based on their likelihood. The authors define three MC moves: node replacement, root addition/removal, and elementary tree replacement, which together enable construction of expression trees while maintaining detailed balance, ensuring proper sampling of the posterior.  

Specifically, the probability of some model given some data is defined as:

\begin{gather}
    p(f_i|D) = \frac{1}{Z}\int_{\Theta_i} d\theta_i p(D|f_i, \theta_i) p(\theta_i|f_i) p(f_i) = \frac{\exp[-\mathcal{L}(f_i)]}{Z}
\end{gather}
where $Z$ is the probability of the dataset $p(D)$, $\Theta_i$ is the space of possible values for parameters $\theta_i$ and $\mathcal{L}$ is the description length of the model.

A central idea in BSR is the inclusion of a prior to emphasize expressions that are \emph{a priori} more likely than others, regardless of the data. Guimera, et al. based their prior off of a corpus of 4080 mathematical expressions collected from Wikipedia (from the ``list of scientific equations named after people"), and assigned the prior likelihood of a function $p(f_i)$ (and its "energy" EP) using the counts of each unique operator ($n_o$) in the corpus, by fitting parameters $\alpha$ and $\beta$ like so:
\begin{gather} \label{equ:p(f_i)}
    \text{EP} = -\log(p(f_i)) = \sum_{o \in O}\big[\alpha_on_o(f_i) + \beta_on_o^2(f_i)\big]
\end{gather}

While this method leads to a distribution of expressions that resembles the corpus when run with no data, $p(f_i)$ can also be set to a constant value so that there is no bias based on operators present in the search process. For our problem, we crafted a prior especially for adsorption thermodynamics (see details in Methods).

\section{Methods}

\renewcommand{\algorithmicrequire}{\textbf{Input:}}
\renewcommand{\algorithmicensure}{\textbf{Output:}}

\subsection{Checking Thermodynamic Constraints}
Three constraint checking functions for the thermodynamic constraints described in Section \ref{sec:OriginalThermoConstraints}) were developed using the Python library SymPy, an open-source computer algebra system\cite{10.7717/peerj-cs.103}. Each function returns either true or false, depending on if its constraint is met or not (if a time limit is exceeded, the constraint is returned as false). For both PySR and BSR, we found that hard constraints (rejecting every expression that fails any constraint) severely hinder the search process, cutting off intermediate expressions between better expressions that may also pass the constraints. Consequently, we impose these as ``soft'' constraints, penalizing expressions for constraint violation, without outright rejecting them. \cite{kronbergerShapeconstrainedSymbolicRegression2022} and \cite{haiderShapeconstrainedMultiobjectiveGenetic2023} also found soft constraints to be more effective than hard constraints. 
This approach (as implemented in PySR) is detailed in algorithm \ref{alg:constraints}.

Constraints 1 and 2 could be checked using SymPy's limit and derivative functionality, but Constraint 3 was more challenging. Though SymPy can check if an expression is strictly increasing in a given range, the check for monotonicity returns false if any change in curvature (critical point) exists for the expression -- thus preventing functions such as $x^3$ from being considered monotonically non-decreasing. 
To allow for zero slope, we implemented a custom monotonic non-decreasing check function (see alg. \ref{alg:monotonic}). Instead of just checking the slope in one range, it checks the ranges between all critical points (as well as to the start and end of the original range in question). 

We hypothesize that the ``equation space'' explored by SR includes accurate, but not thermodynamically consistent expressions that can be rejected through the incorporation of background knowledge, guiding the search to more theory-informed expressions.  

\subsection{PySR Modifications}
In PySR, each member in a population has a score to be minimized, which combines the loss and complexity (defined by total nodes in the expression tree). When a thermodynamic constraint is violated, we multiply the loss function by a penalty, raising the score and making the expression less fit. This allows any number of constraints to be checked in any order (as multiplication is commutative), and confers larger penalties to expressions that violate multiple constraints.

\begin{gather}
    \text{Loss: } L = \ell_2^R * \prod_{i=1,2,3} c_{i}^{\delta_i} \text{ where } \delta_i = 
    \left\{
    \begin{array}{lr}
        1 & \text{if constraint $i$ passed} \\
        0 & \text{if constraint $i$ failed}
    \end{array} \right\} \\ 
    \text{Member Score: } S = L + n_{nodes} * c_{l}
\end{gather}

The above equations detail how the loss and score are calculated in PySR.
$\ell_2^R$ is the L2 norm, $c_i$ is the penalty for constraint $i$, $\delta_i$ indicates if constraint $i$ is passed and $c_l$ is the penalty for the length / complexity of an expression.


PySR also has the option to take any operators defined in Julia or Python, including custom user-defined operators.  For this work only the operators $+$, $-$, $*$ and $\div$ were used to manage the size of the search space.
Expressions written in their canonical form may use other operators such as exponents but these are only due to simplification of generated expressions.

\subsection{BMS Modifications}

The prior used in the Bayesian Machine Scientist code by Guimera et al. \cite{guimera_bayesian_2020} incorporates ``background knowledge'' in its equation search by considering mathematical operation frequency among named equations in Wikipedia. The authors found this to be helpful for searching for general scientific equations, but we aim to color this background knowledge according to our domain of inquiry.
We consider the thermodynamic constraints described above to be our ``prior knowledge,'' and construct the following expression:

\begin{gather}
    \text{EP} = \sum_{o \in O}\big[c_{ops}n_o(f_i)\big] + \sum_{i=1,2}c_i*\delta_i \text{ where } \delta_i
    = \left\{ \begin{array}{lr}
        1 & \text{if constraint $i$ passed} \\
        0 & \text{if constraint $i$ failed} 
    \end{array} \right\}
\end{gather}

where $c_{ops}$ is the constraint penalty for operators (analogous to the parsimony parameter in PySR), and $n_{o}$ is the count of each operator in expression $f_i$.
This expression directly replaces Eq.~\ref{equ:p(f_i)}, changing the prior distribution.  Note that we checked all three constraints with PySR, and only the first two constraints with BSR (omitting the monotonic non-decreasing check).

\section{Results}

\subsection{Datasets}



\begin{figure}[H]
    \centering
    \includegraphics[width=0.45\textwidth]{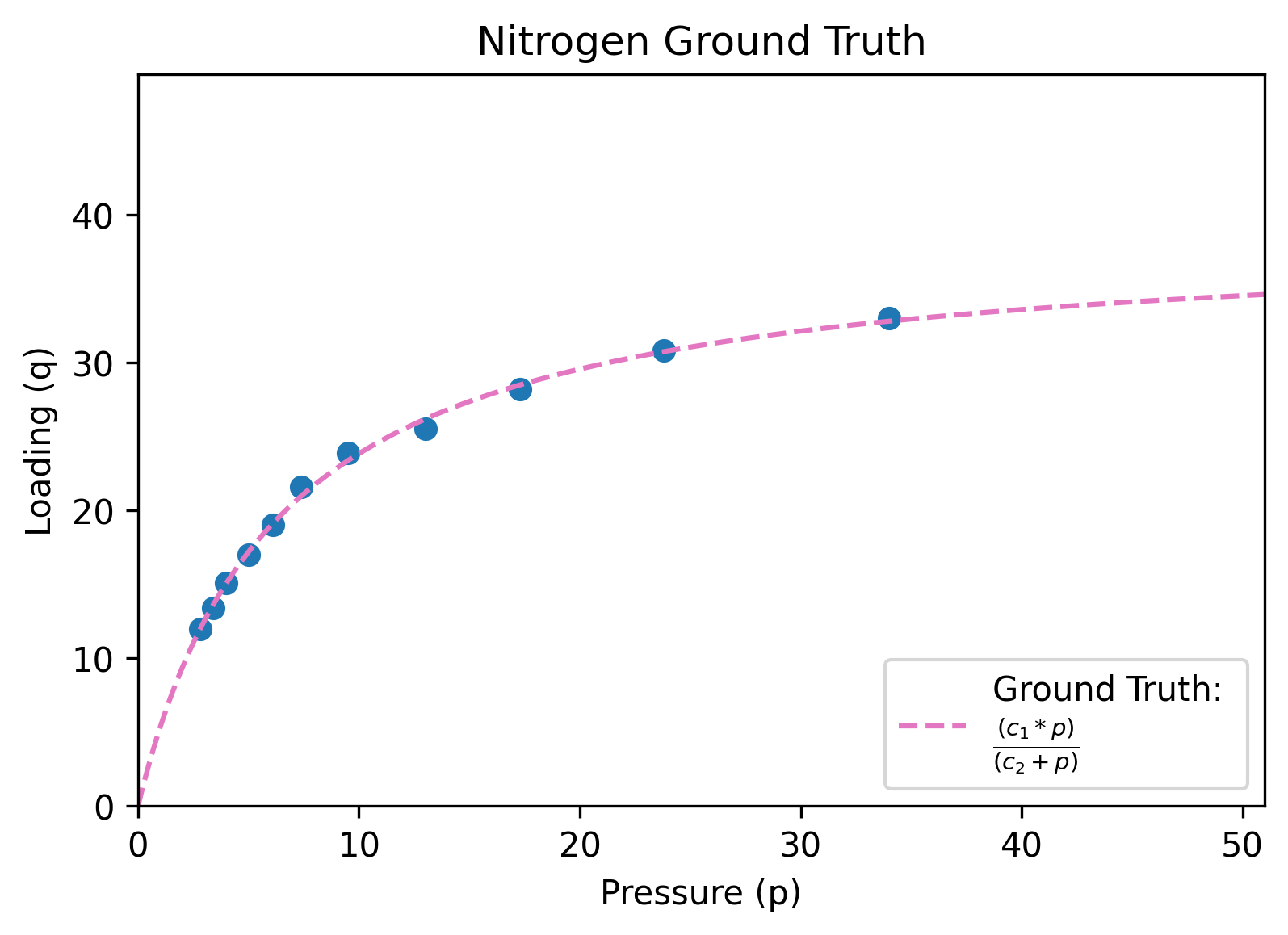}
    \includegraphics[width=0.45\textwidth]{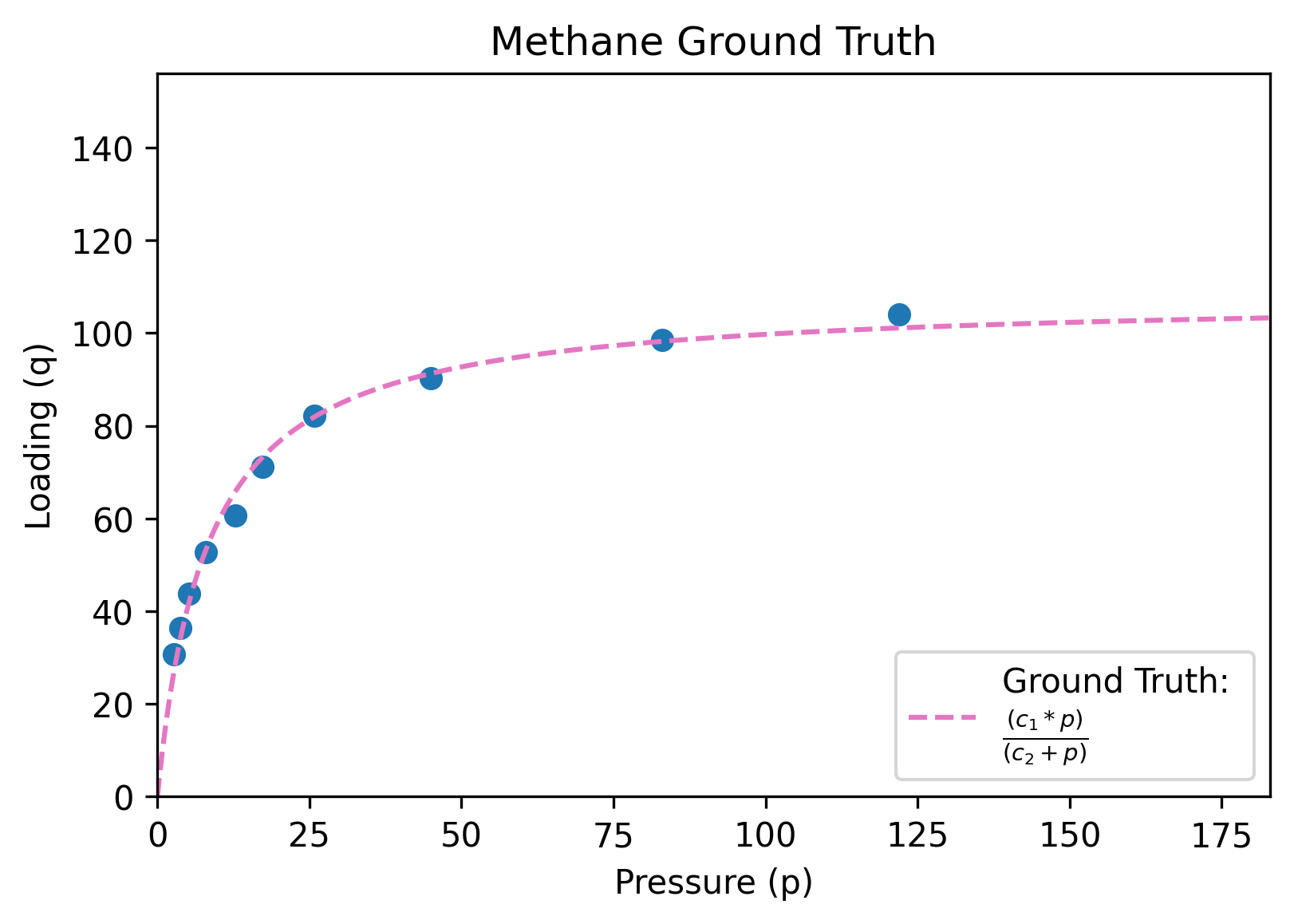}
    \includegraphics[width=0.45\textwidth]{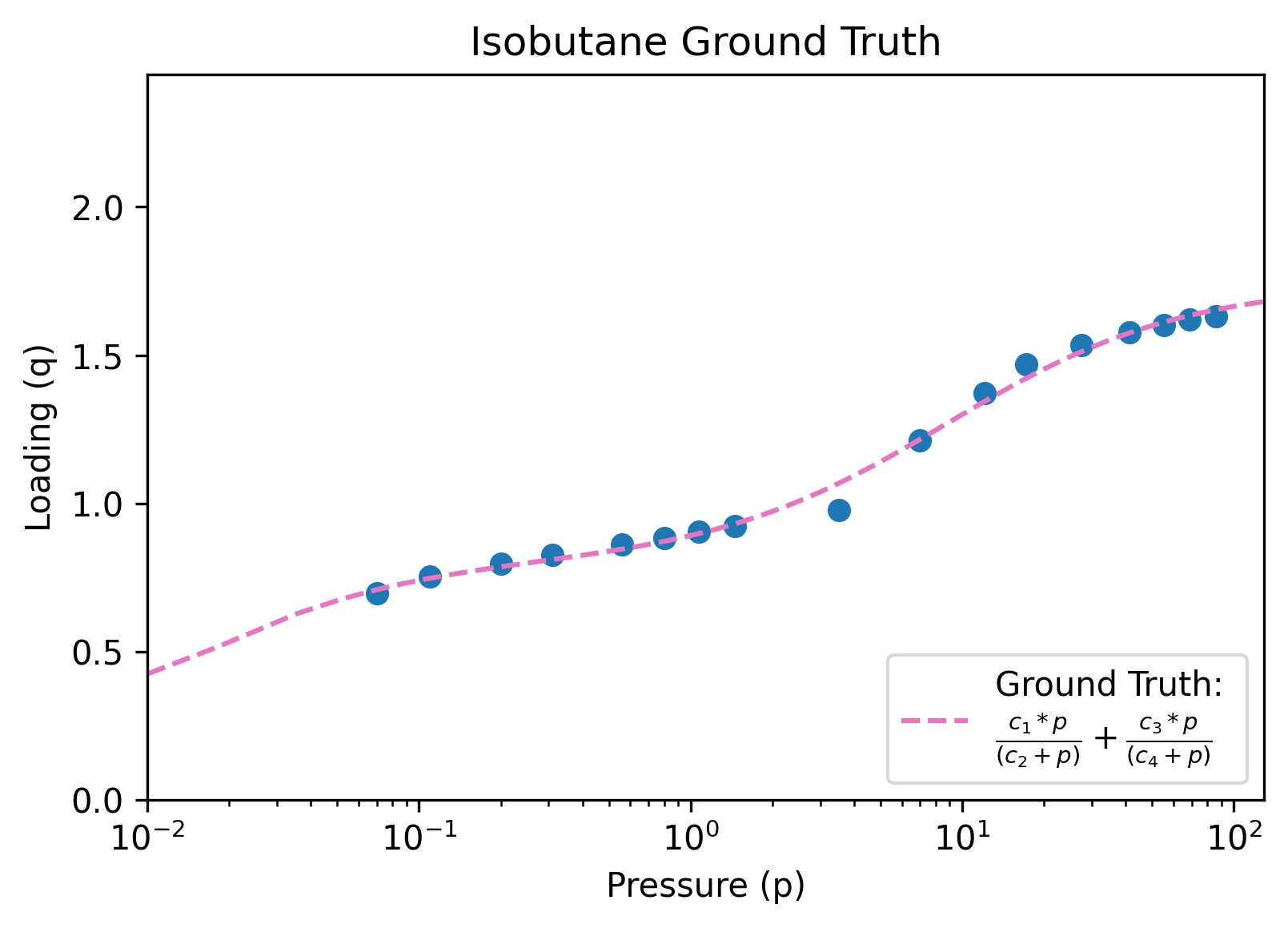}
    \includegraphics[width=0.45\textwidth]{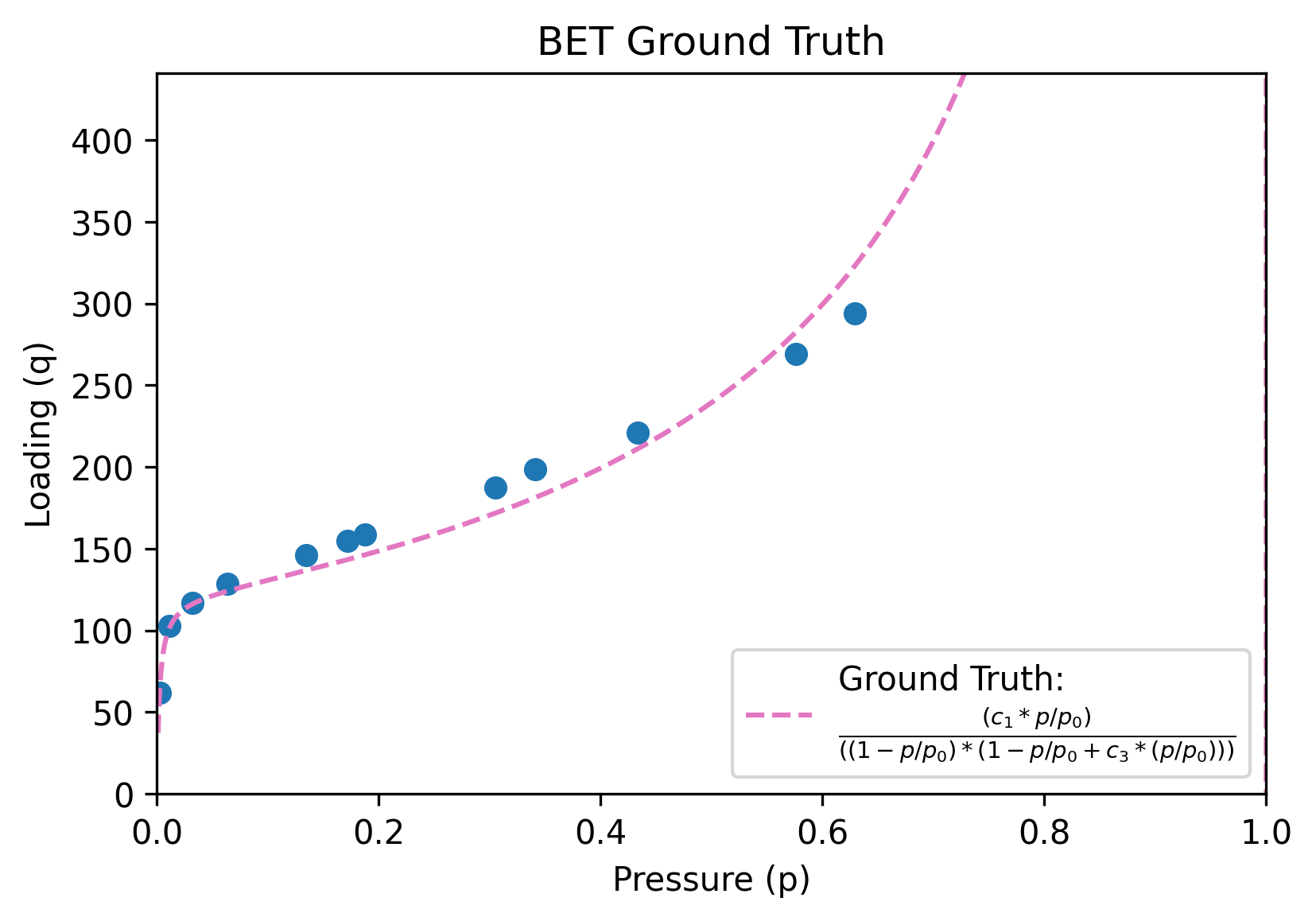}
    \caption{Each dataset and corresponding ground truth model with constants fit using SciPy.  Isobutane is shown with log scaled pressure so that the two separate curves are visible.  BET is shown with pressure increasing to 1 so the asymptote is visible.}
    \label{fig:allGT}
\end{figure}

To examine the effects of adding constraints to SR during a search, four experimental adsorption datasets were identified: adsorption of nitrogen and methane on mica \cite{langmuir_adsorption_1918}, adsorption of isobutane in silicalite,  \cite{vlugt_adsorption_1998} and adsorption of nitrogen on Fe-Al$_2$0$_3$ catalyst \cite{brunauer_adsorption_1938}.
The first and second datasets come from the landmark paper introducing the Langmuir isotherm model \cite{langmuir_adsorption_1918}.
This model assumes there are discrete loading "sites" that do not interact with each other, and that each site can either be occupied or not.
The isobutane dataset is well-described by a dual-site Langmuir model which has two unique types of sites.
The fourth dataset (referred to as the BET dataset) was used by the authors of BET theory to support their model for multilayer adsorption. These data and the respective ground truth model fits are shown in Fig.~\ref{fig:allGT}. 

\subsection{Langmuir Datasets}




The main results of this work are shown in two plot types. 
The left column contains Pareto fronts which show the best expressions based on complexity and accuracy. In these, the horizontal axis shows increasing complexity (defined as the total number of nodes in an expression tree), and the vertical axis shows loss, which is logarithmically scaled so the trend of the Pareto front is more apparent. 
The best expression at each complexity is taken from each of 8 runs (gray curves), with the overall Pareto front shown in orange.
The ``ground truth" expression for each dataset is also shown in the form it would likely be expressed by SR, along with loss found using fit constants.
The right column of each figure shows the dataset and select expressions from the overall Pareto front for that test. 
Only some expressions are shown so plots remain readable and because expressions longer than the ground truth are usually overfit and overlay the ground truth expression too closely for distinction. 
The ground truth is plotted with a dotted line so that expressions with similar accuracy can still be seen. Plotting the generated expressions on the data helps to illustrate how they may or may not follow the thermodynamic constraints and how similar they are to the ground truth.

\begin{figure}[H]
    \centering
    \begin{subfigure}[b]{0.45\textwidth}
        \caption{}
        \includegraphics[trim={0 0 0 0.6cm}, width=\textwidth]{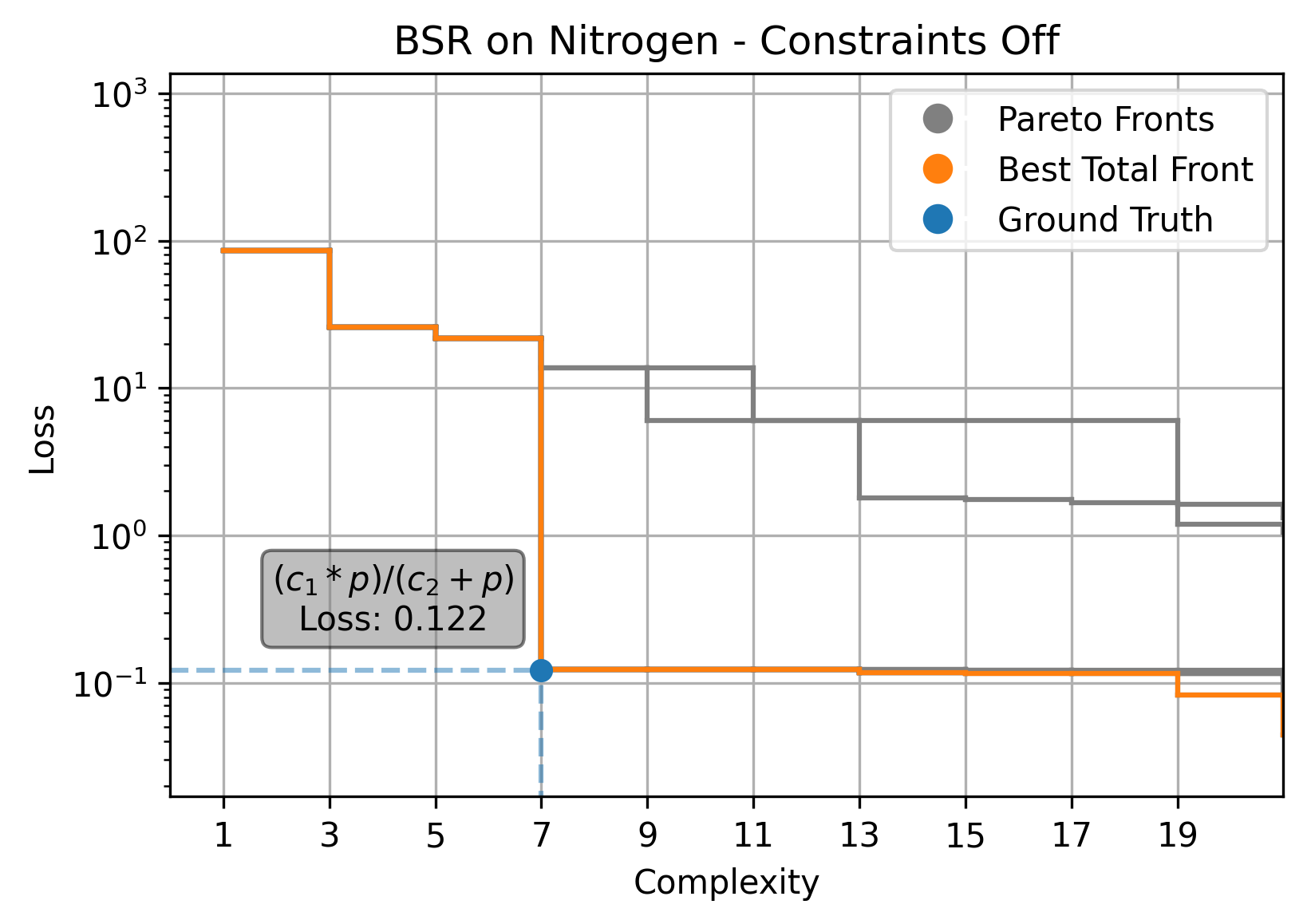}
        \label{fig:BSRNitrogenOffPareto}
    \end{subfigure}%
    \hfill
    \begin{subfigure}[b]{0.45\textwidth}
        \caption{}
        \includegraphics[trim={0 0 0 0.6cm}, width=\textwidth]{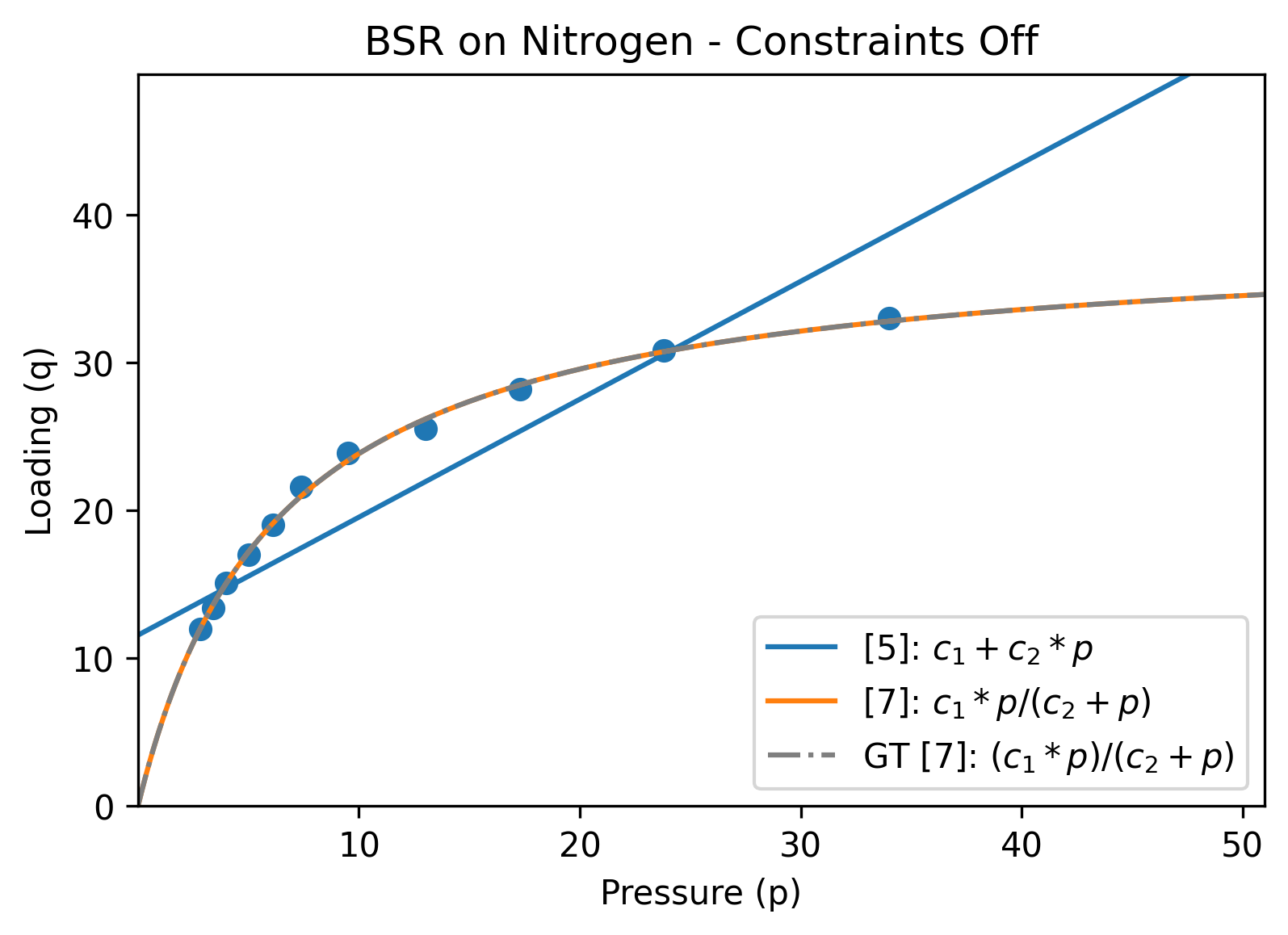}
        \label{fig:BSRNitrogenOffIsotherm}
    \end{subfigure}%
    \vspace{-2\baselineskip}
    \begin{subfigure}[b]{0.45\textwidth}
        \caption{}
        \includegraphics[trim={0 0 0 0.6cm}, width=\textwidth]{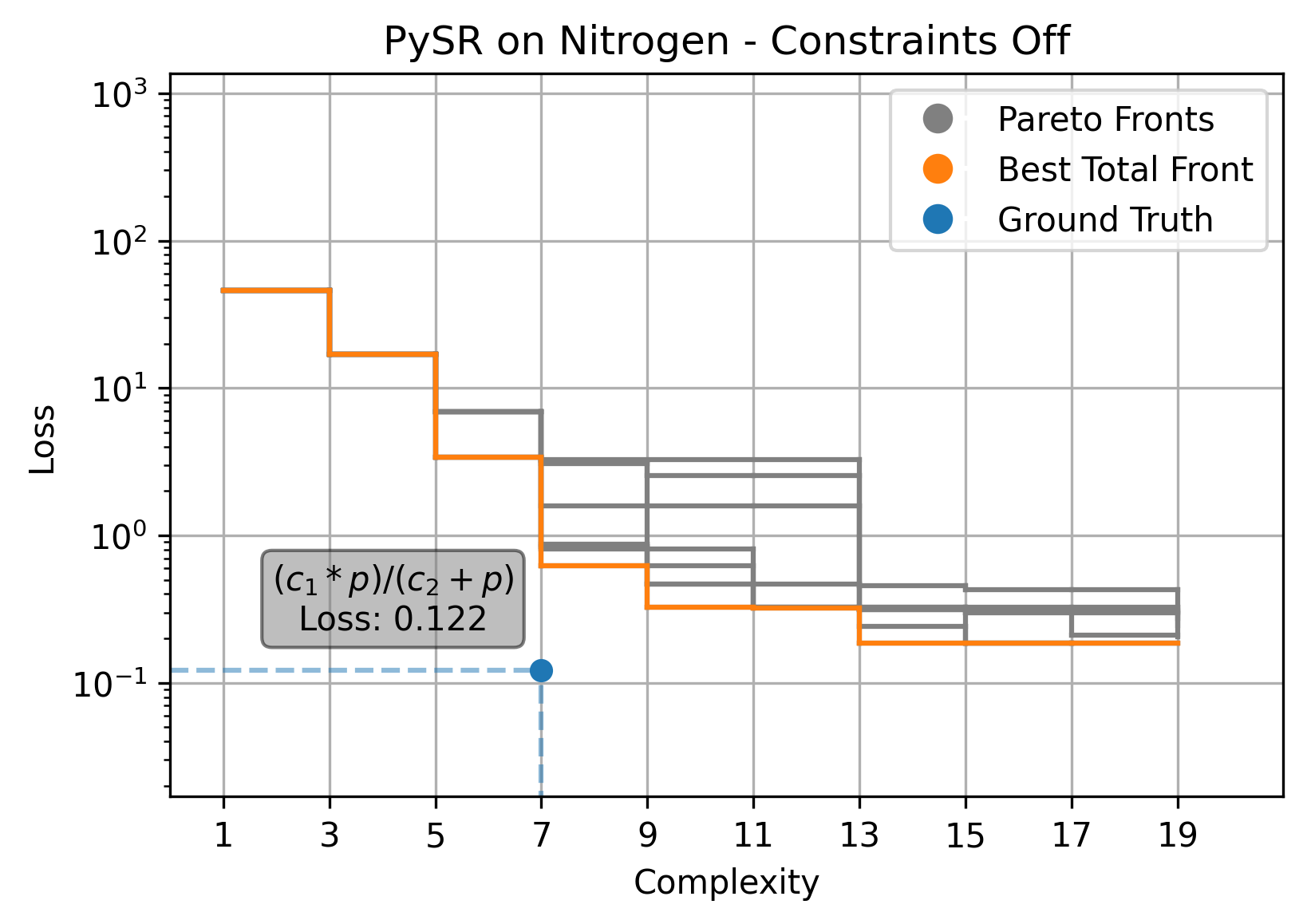}
        \label{fig:PySRNitrogenOff}
    \end{subfigure}%
    \hfill
    \begin{subfigure}[b]{0.45\textwidth}
        \caption{}
        \includegraphics[trim={0 0 0 0.6cm}, width=\textwidth]{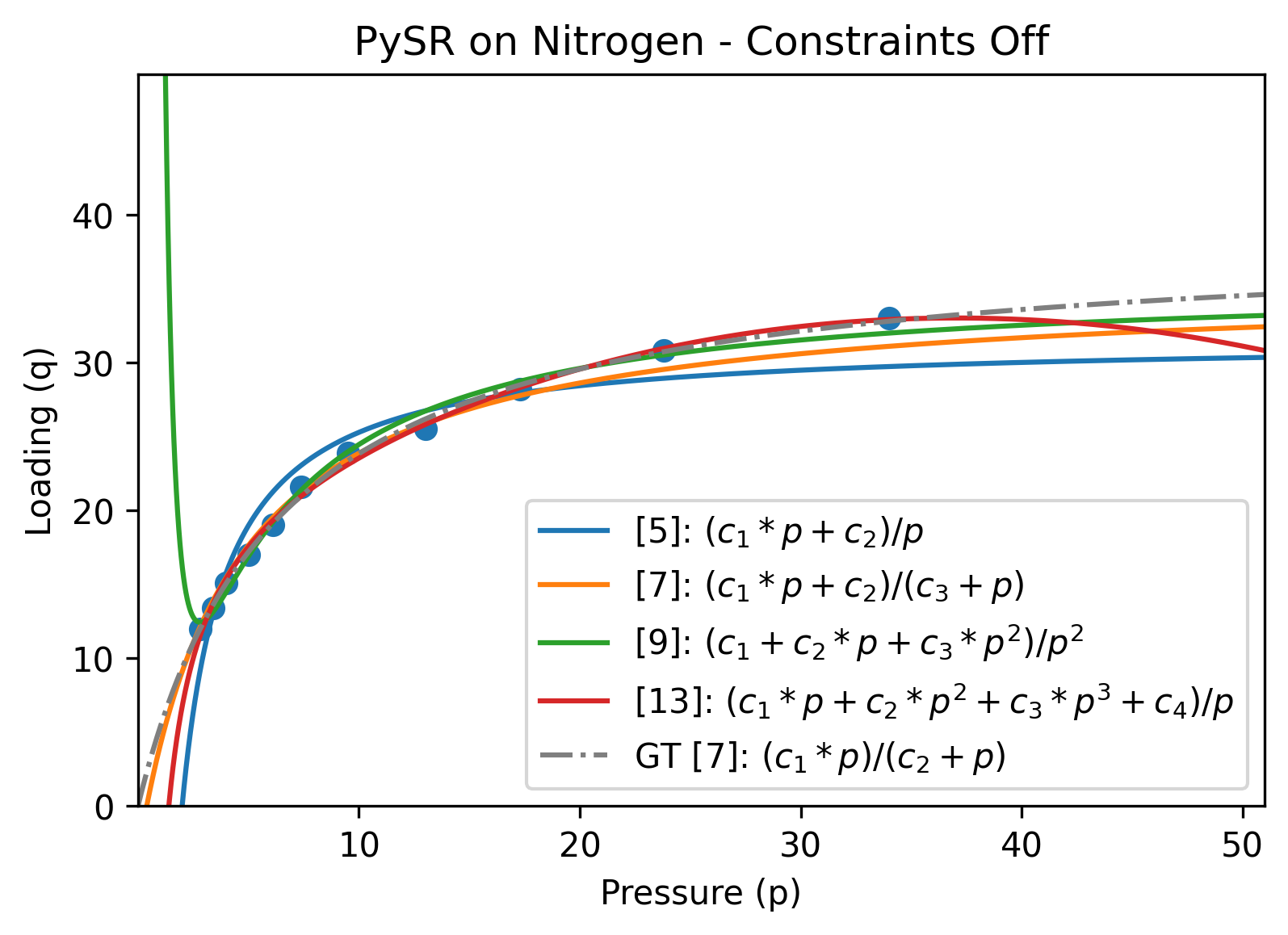}
        \label{fig:PySRNitrogenOffCorners}
    \end{subfigure}%
    \vspace{-2\baselineskip}
    \begin{subfigure}[b]{0.45\textwidth}
        \caption{}
        \includegraphics[trim={0 0 0 0.6cm}, width=\textwidth]{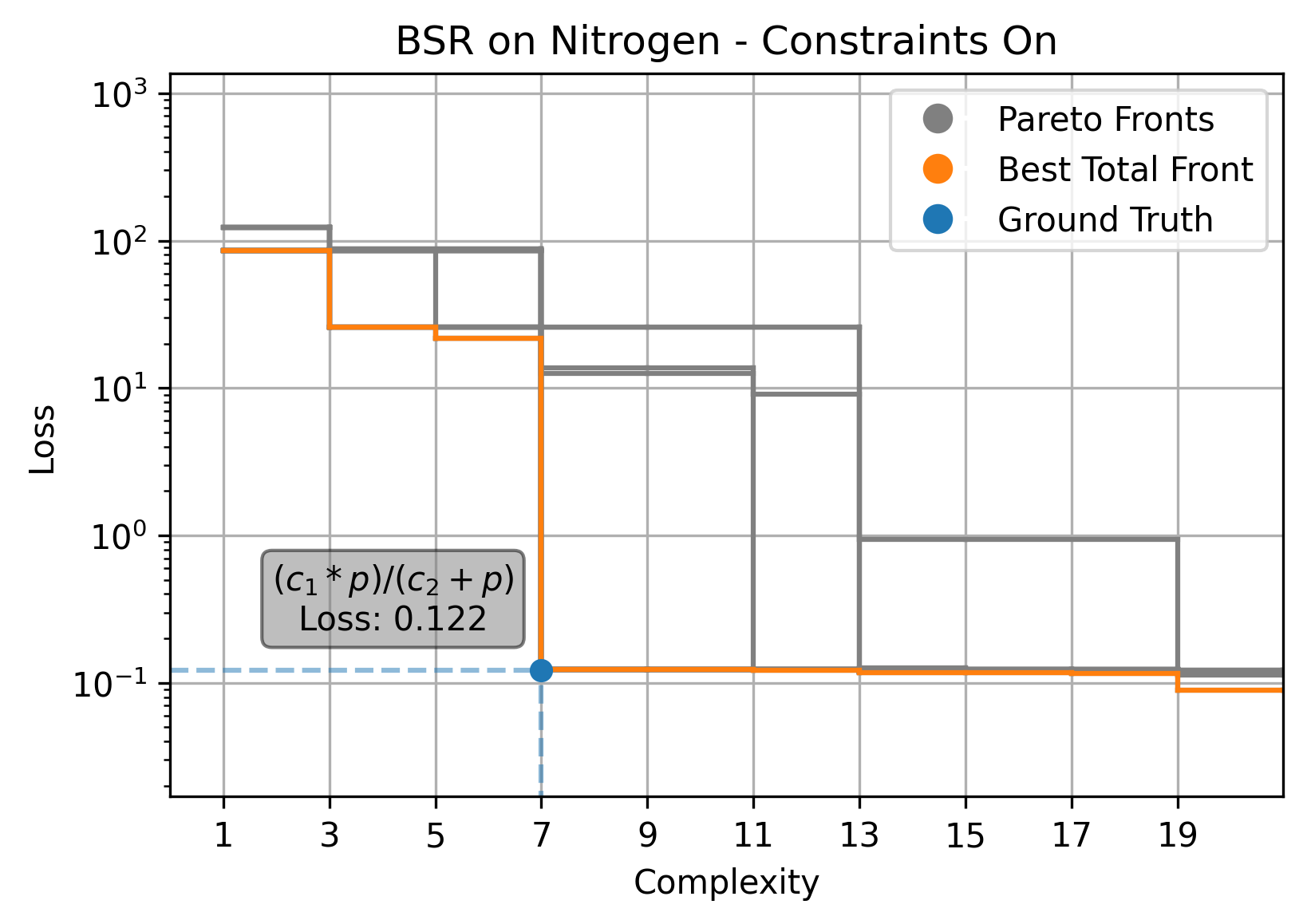}
        \label{fig:BSRNitrogenOnPareto}
    \end{subfigure}%
    \hfill
    \begin{subfigure}[b]{0.45\textwidth}
        \caption{}
        \includegraphics[trim={0 0 0 0.6cm}, width=\textwidth]{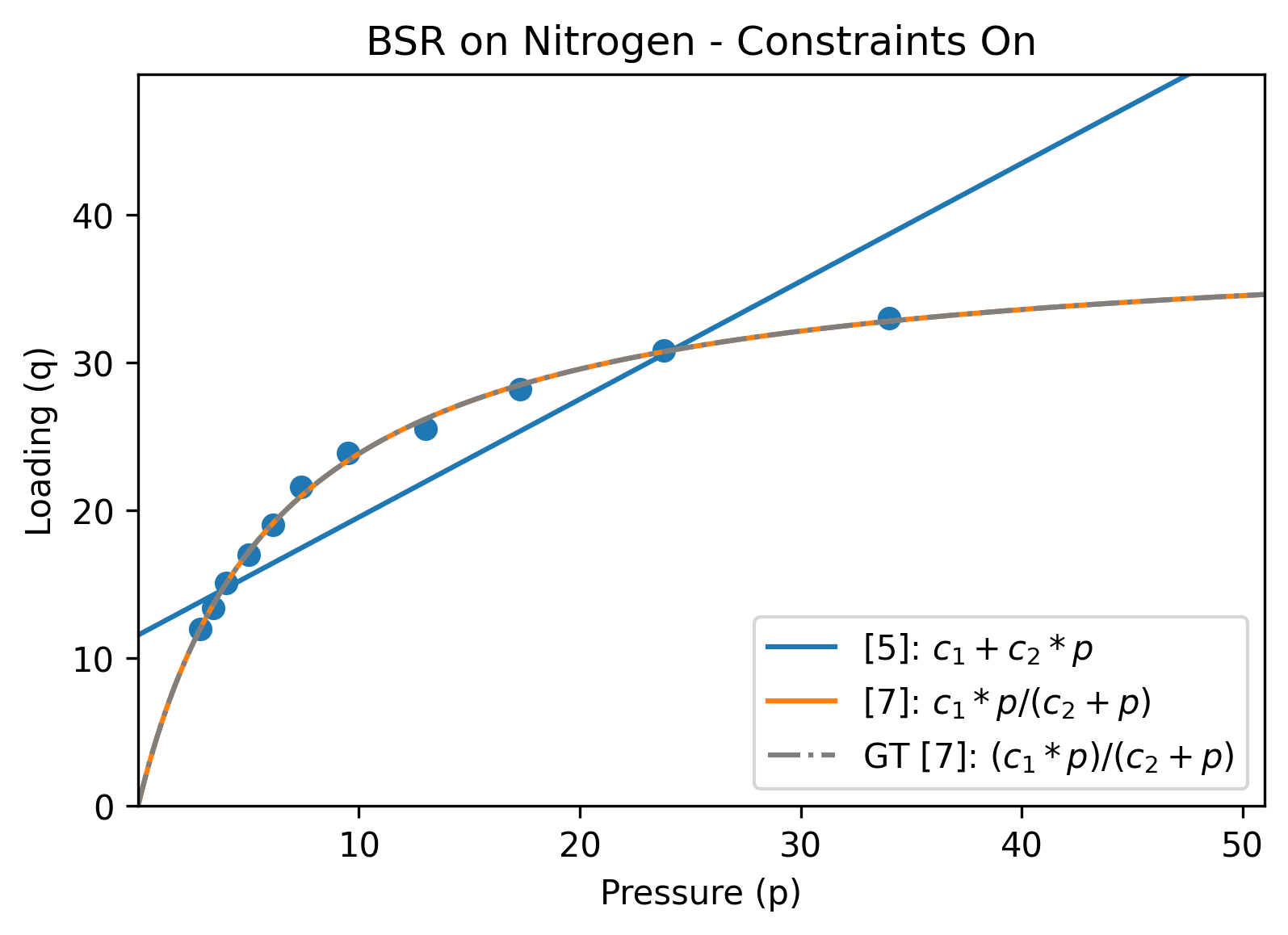}
        \label{fig:BSRNitrogenOnIsotherm}
    \end{subfigure}%
    \vspace{-2\baselineskip}
    \begin{subfigure}[b]{0.45\textwidth}
        \caption{}
        \includegraphics[trim={0 0 0 0.6cm}, width=\textwidth]{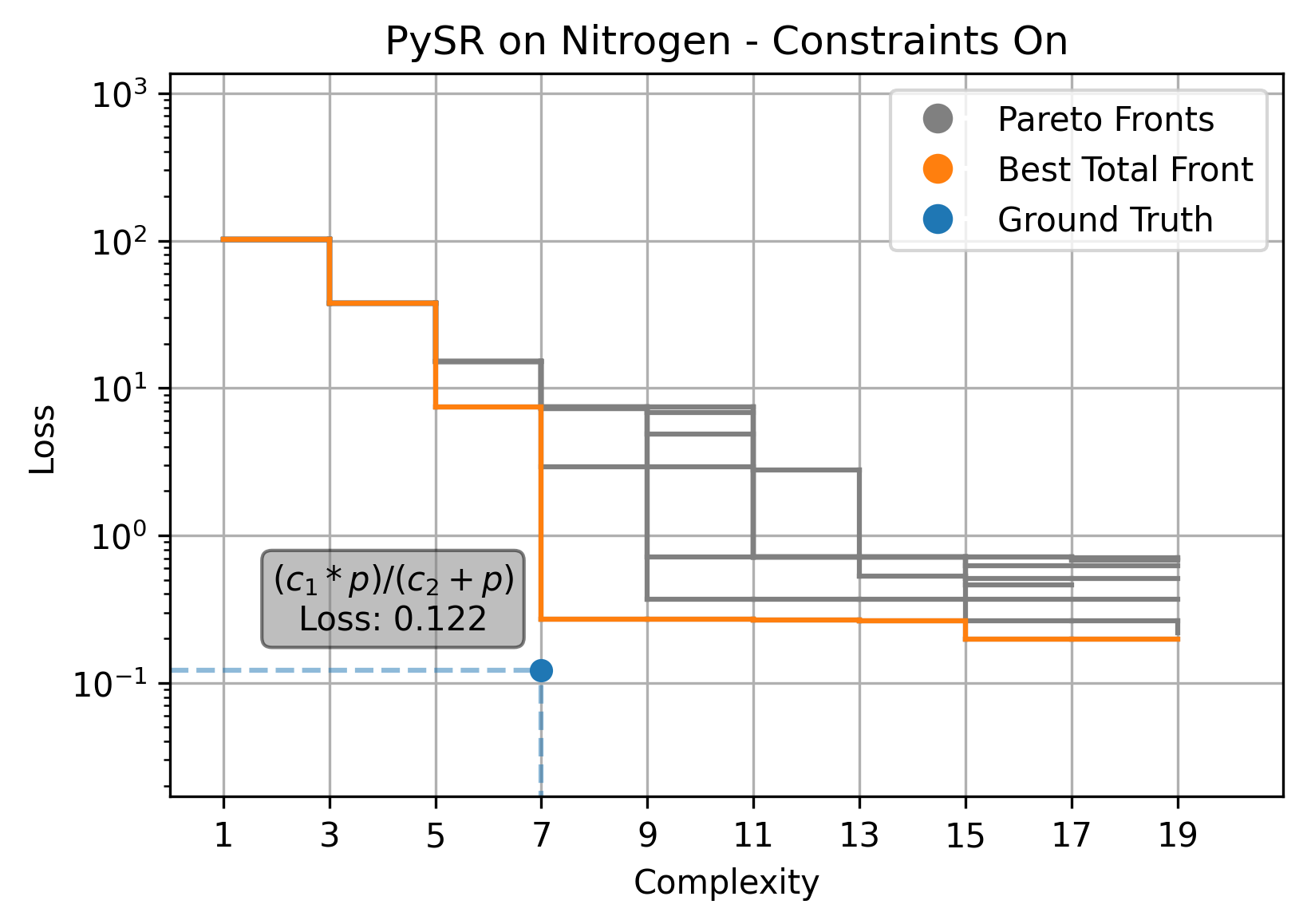}
        \label{fig:PySRNitrogenOnPareto}
    \end{subfigure}%
    \hfill
    \begin{subfigure}[b]{0.45\textwidth}
        \caption{}
        \includegraphics[trim={0 0 0 0.6cm}, width=\textwidth]{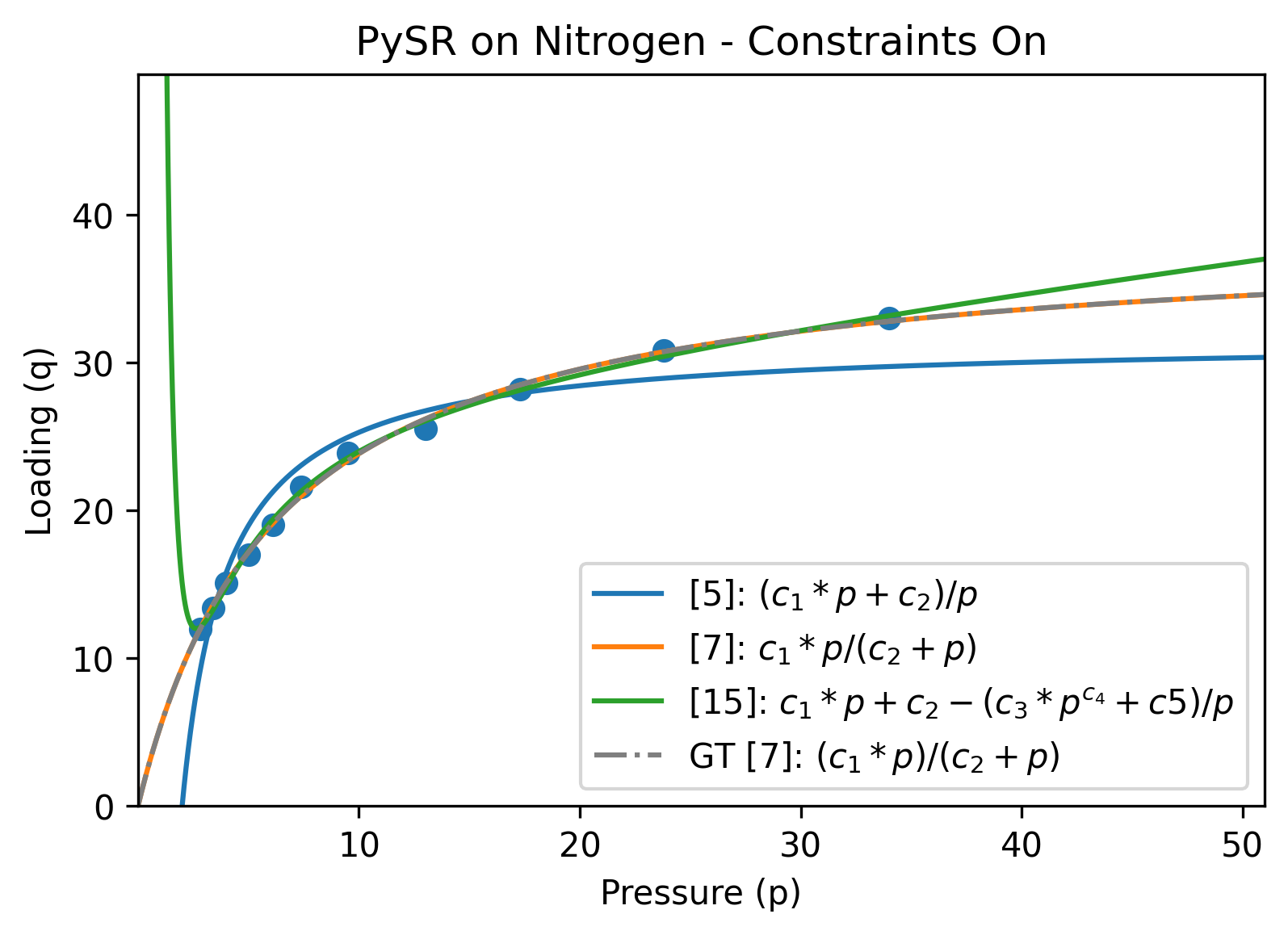}
        \label{fig:PySRNitrogenOnIsotherm}
    \end{subfigure}%
    \vspace{-1\baselineskip}
    \caption{BSR and PySR on the Nitrogen dataset.  The left column shows combined Pareto fronts across 8 runs and the right column shows interesting isotherms found at the defining corners of those Pareto fronts.  The constraints are disabled in the top four subplots and enabled in the bottom four.  The rows alternate between BSR and PySR.}
    \label{fig:Nitrogen}
\end{figure}

Figure ~\ref{fig:Nitrogen} shows the results from both SR algorithms with constraints on and off on the Langmuir nitrogen dataset. 
The first and second rows show BSR and PySR respectively with constraints off and clearly show that BSR finds the ground truth while PySR does not.  The expression that defines the corner at complexity 7 in the BSR Pareto front plot (Fig. \ref{fig:BSRNitrogenOffPareto}) is indistinguishable from the ground truth (both written mathematically and drawn on the data) when viewed in the isotherm plot (Fig. \ref{fig:BSRNitrogenOffIsotherm}).
The BSR plot (Fig. \ref{fig:BSRNitrogenOffPareto}) has a much larger variance in terms of best Pareto fronts across 8 runs (as shown by the grey lines) than PySR, but this may indicate longer time needed for the algorithm to converge. 
The corresponding isotherm plots (the right column) show how expressions fit the data better as they become more complex, following the general trend of the Pareto fronts. These plots also show how some expressions can fit the data reasonably well while violating the constraints from theory, as is the case in the plot for PySR (Fig. \ref{fig:PySRNitrogenOffCorners}). In fact, only 2.2\% of expressions generated by PySR (without enforcing constraints) pass the first constraint and only 33\% pass the second constraint (Table \ref{tab:PercentPassing}).
Without constraints enforced, BSR finds more consistent expressions than PySR, with 37\% of its expressions passing the first constraint and 67\% passing the second.

When the thermodynamic constraints are enabled, the effect is clearly shown in the Pareto fronts (bottom two rows).  Both SR methods find the ground truth and achieve the same or similar accuracy (accuracy is less for the same expression when the constants were not optimized as thoroughly in the search). Datasets that are well represented by the Langmuir isotherm show the effects of the constraints well because it is typically very accurate as well as being concise.  The isotherm plots show, as before, how the expressions fit the data better as they become more complex but showing anything beyond a complexity of 7 is redundant as the ground truth is discovered and matches the pre-fit ground truth almost perfectly.  The trend of slightly more variation across BSR runs also continues here to some extent and the variation across PySR runs appears roughly similar to with constraints disabled.
Importantly, PySR sees a 5x increase in expressions passing the first constraint (though still only 10\%) and a marginal improvement across the other two constraints (8\% and 13\%). The change is more stark in BSR where twice as many expressions now pass the first constraint (up to 72\%) and a significant portion pass the third constraint (up to 19\% from 0.46\%) even though it was not included in the Bayesian prior.  

\begin{figure}[H]
    \centering
    \begin{subfigure}[b]{0.45\textwidth}
        \caption{}
        \includegraphics[trim={0 0 0 0.6cm}, width=\textwidth]{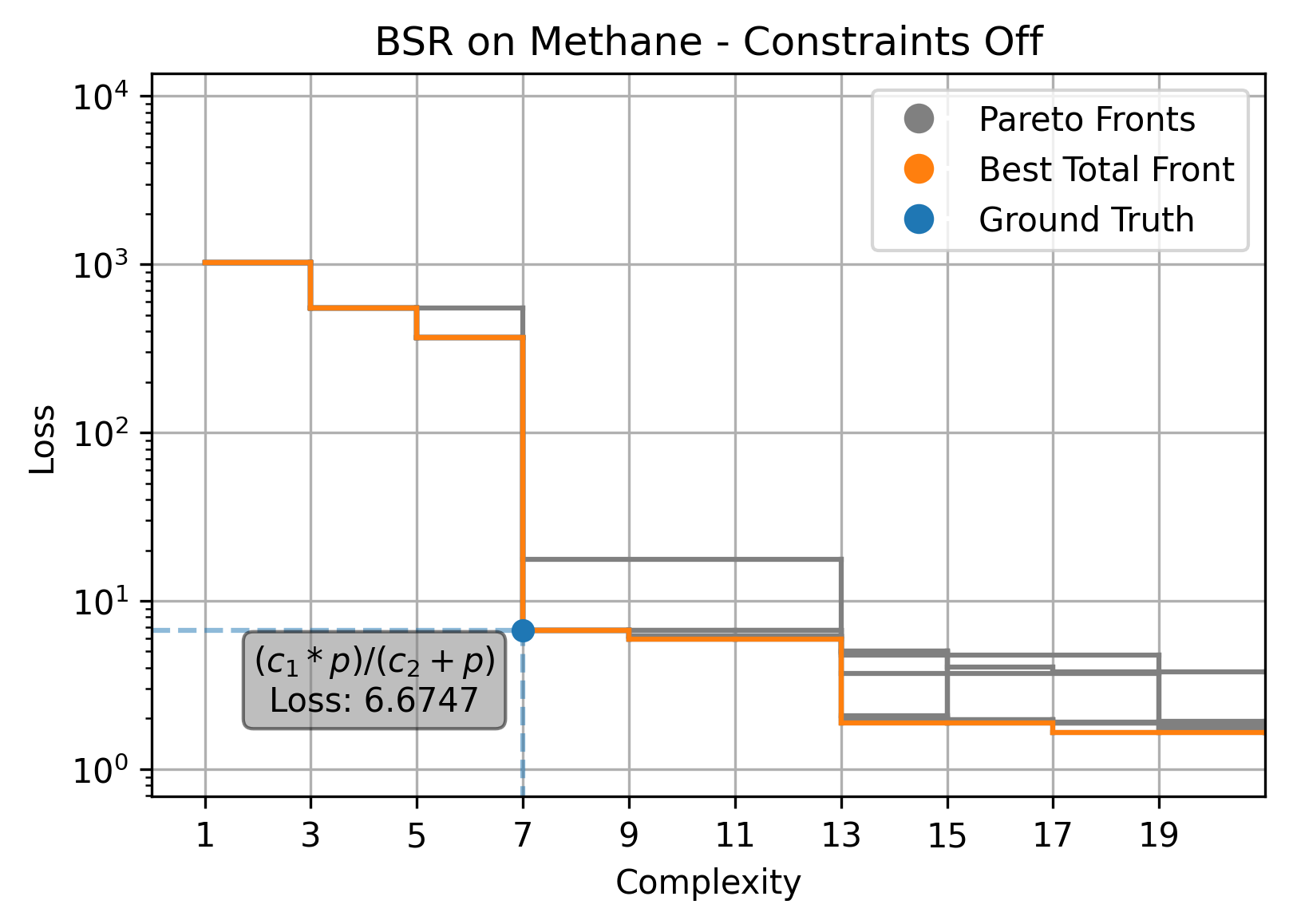}
        \label{fig:BSRMethaneOffPareto}
    \end{subfigure}%
    \hfill
    \begin{subfigure}[b]{0.45\textwidth}
        \caption{}
        \includegraphics[trim={0 0 0 0.6cm}, width=\textwidth]{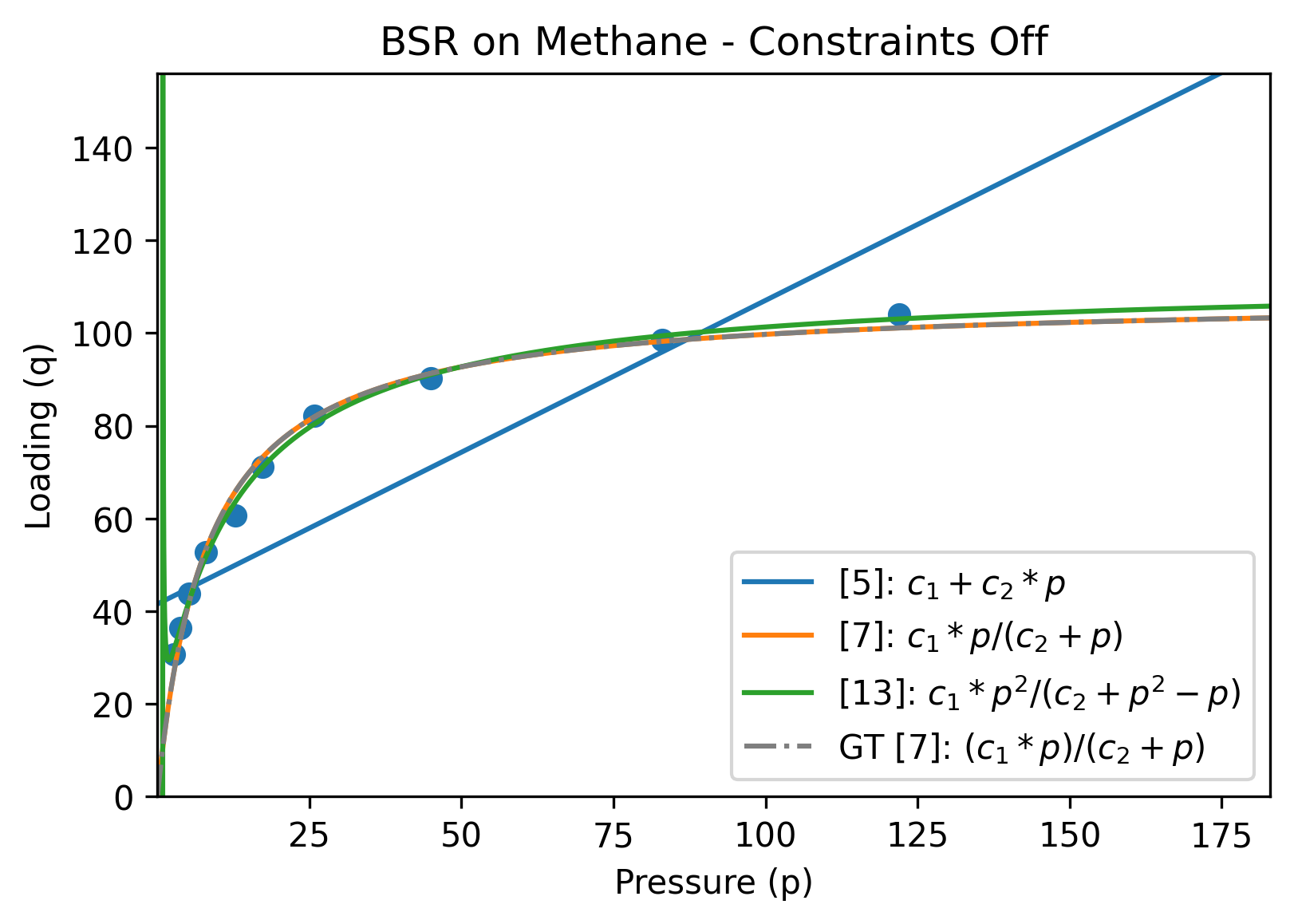}
        \label{fig:BSRMethaneOffIsotherm}
    \end{subfigure}%
    \vspace{-2\baselineskip}
    \begin{subfigure}[b]{0.45\textwidth}
        \caption{}
        \includegraphics[trim={0 0 0 0.6cm}, width=\textwidth]{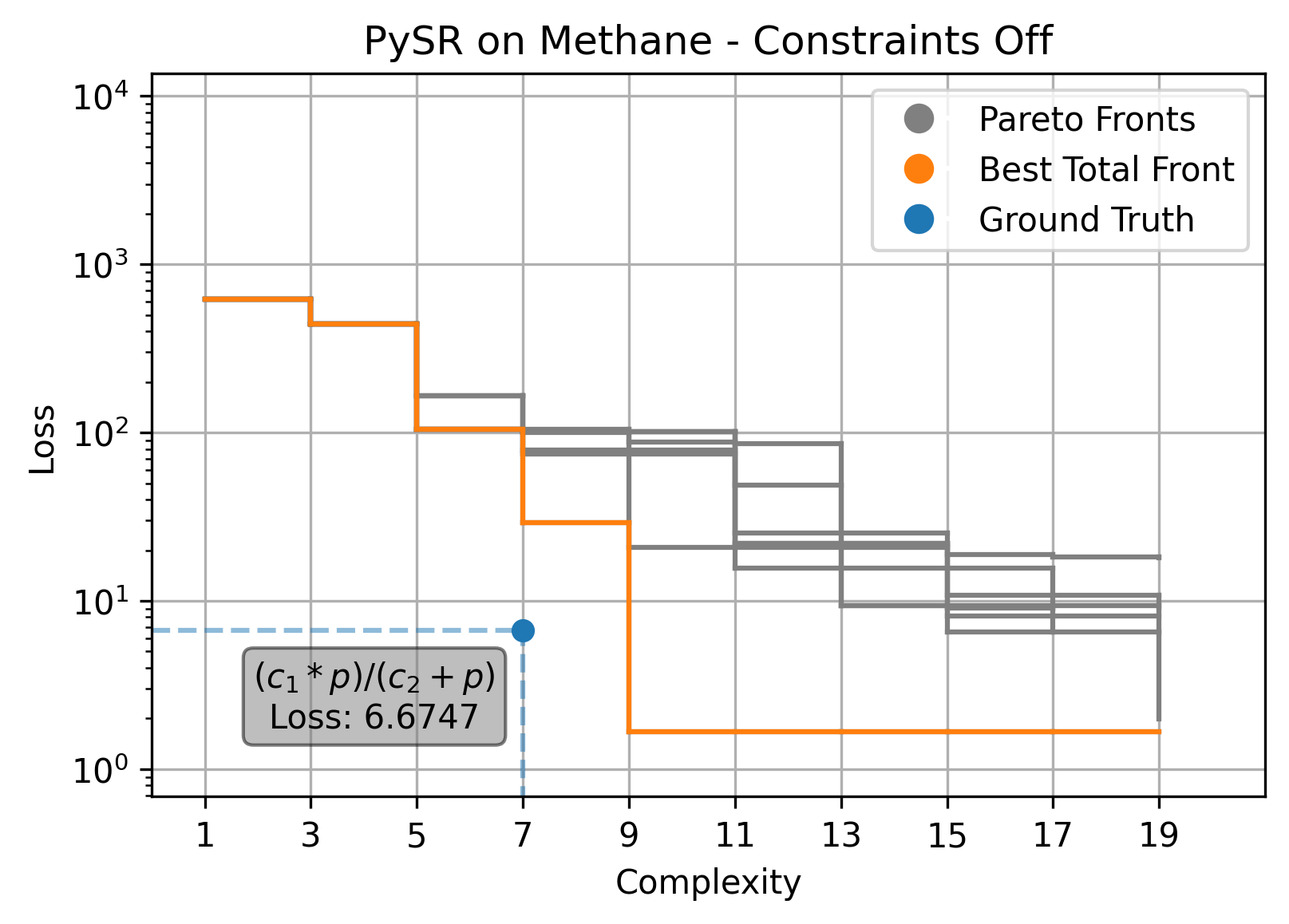}
        \label{fig:PySRMethaneOffPareto}
    \end{subfigure}%
    \hfill
    \begin{subfigure}[b]{0.45\textwidth}
        \caption{}
        \includegraphics[trim={0 0 0 0.6cm}, width=\textwidth]{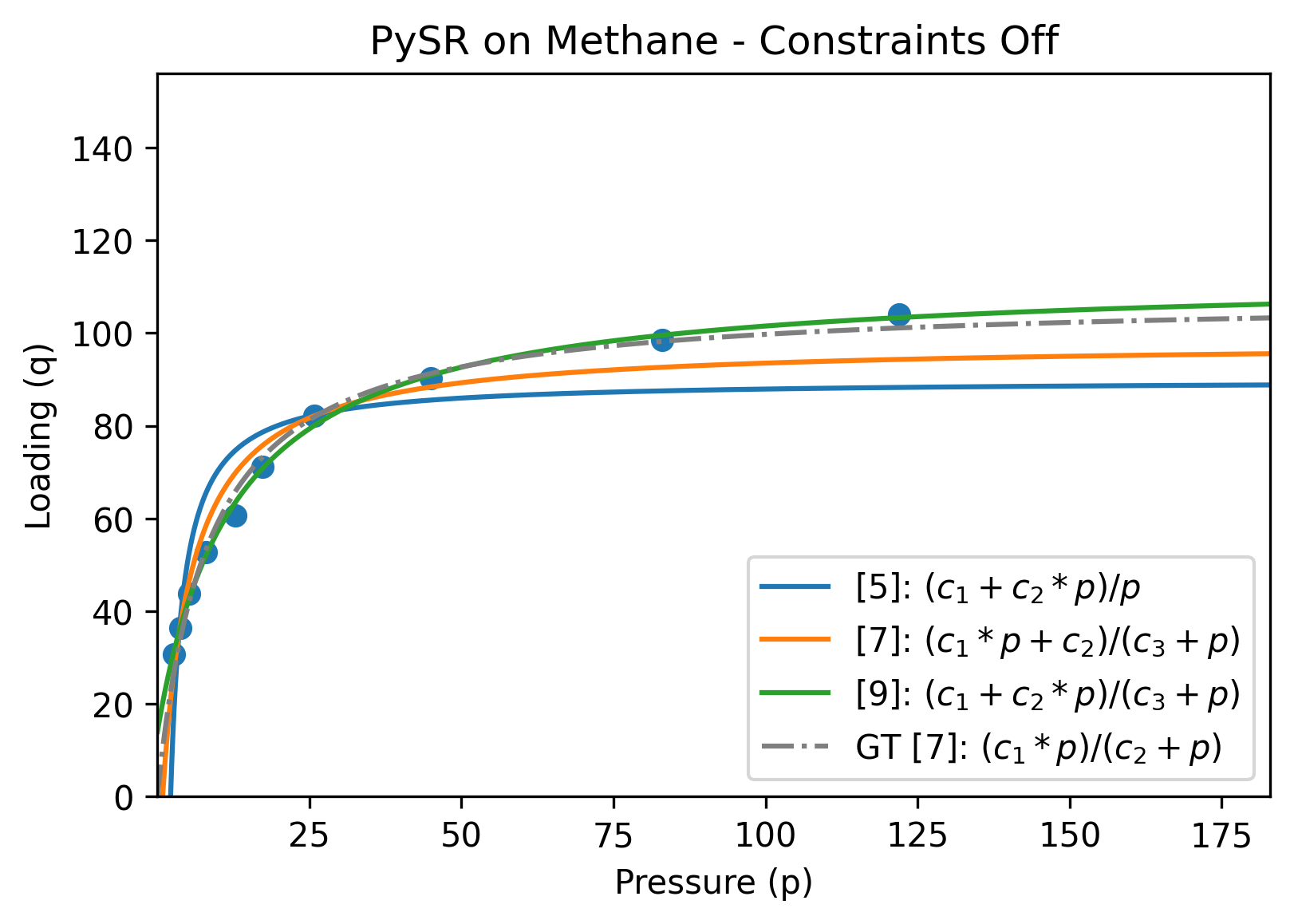}
        \label{fig:PySRMethaneOffIsotherm}
    \end{subfigure}%
    \vspace{-2\baselineskip}
    \begin{subfigure}[b]{0.45\textwidth}
        \caption{}
        \includegraphics[trim={0 0 0 0.6cm}, width=\textwidth]{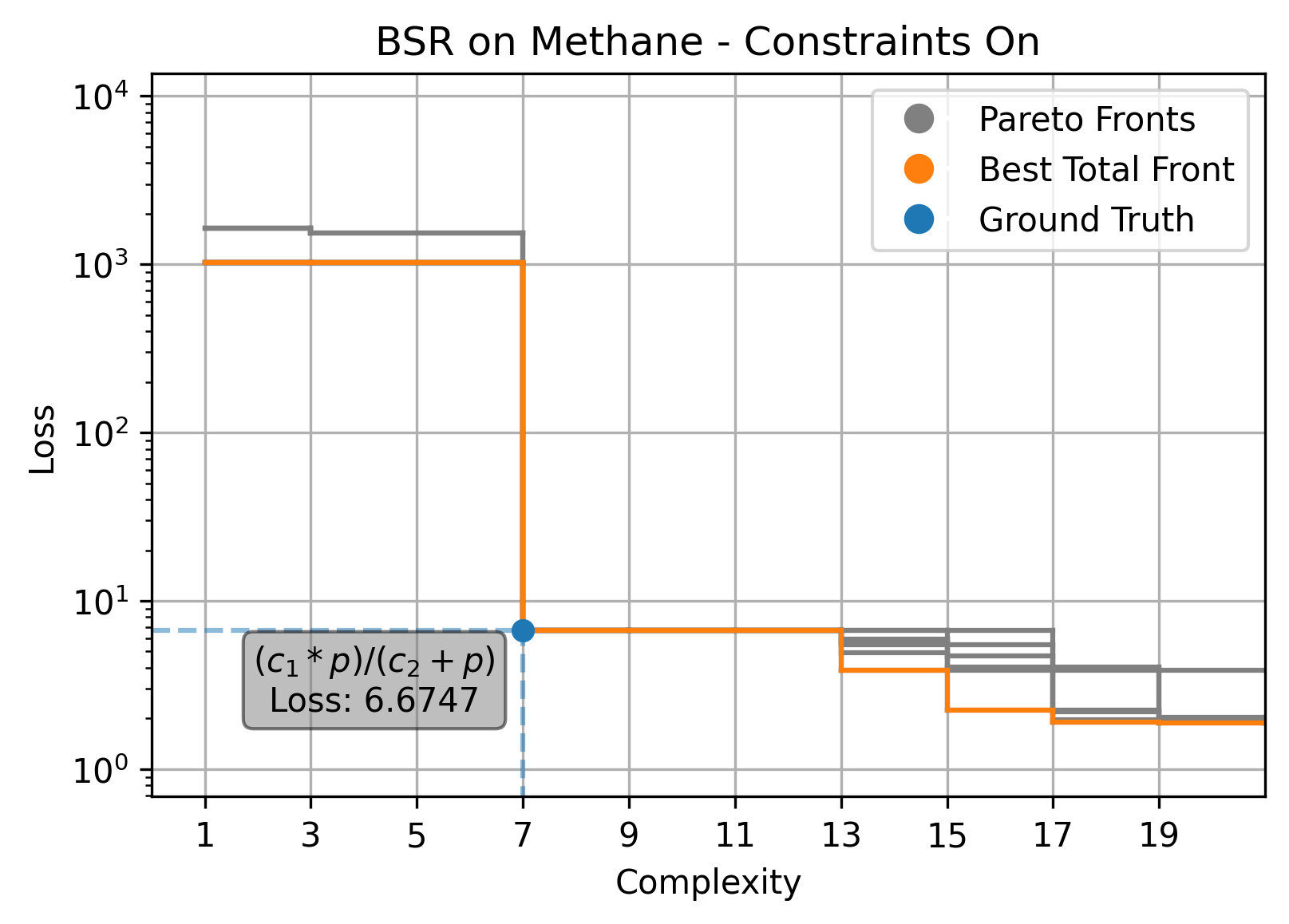}
        \label{fig:BSRMethaneOnPareto}
    \end{subfigure}%
    \hfill
    \begin{subfigure}[b]{0.45\textwidth}
        \caption{}
        \includegraphics[trim={0 0 0 0.6cm}, width=\textwidth]{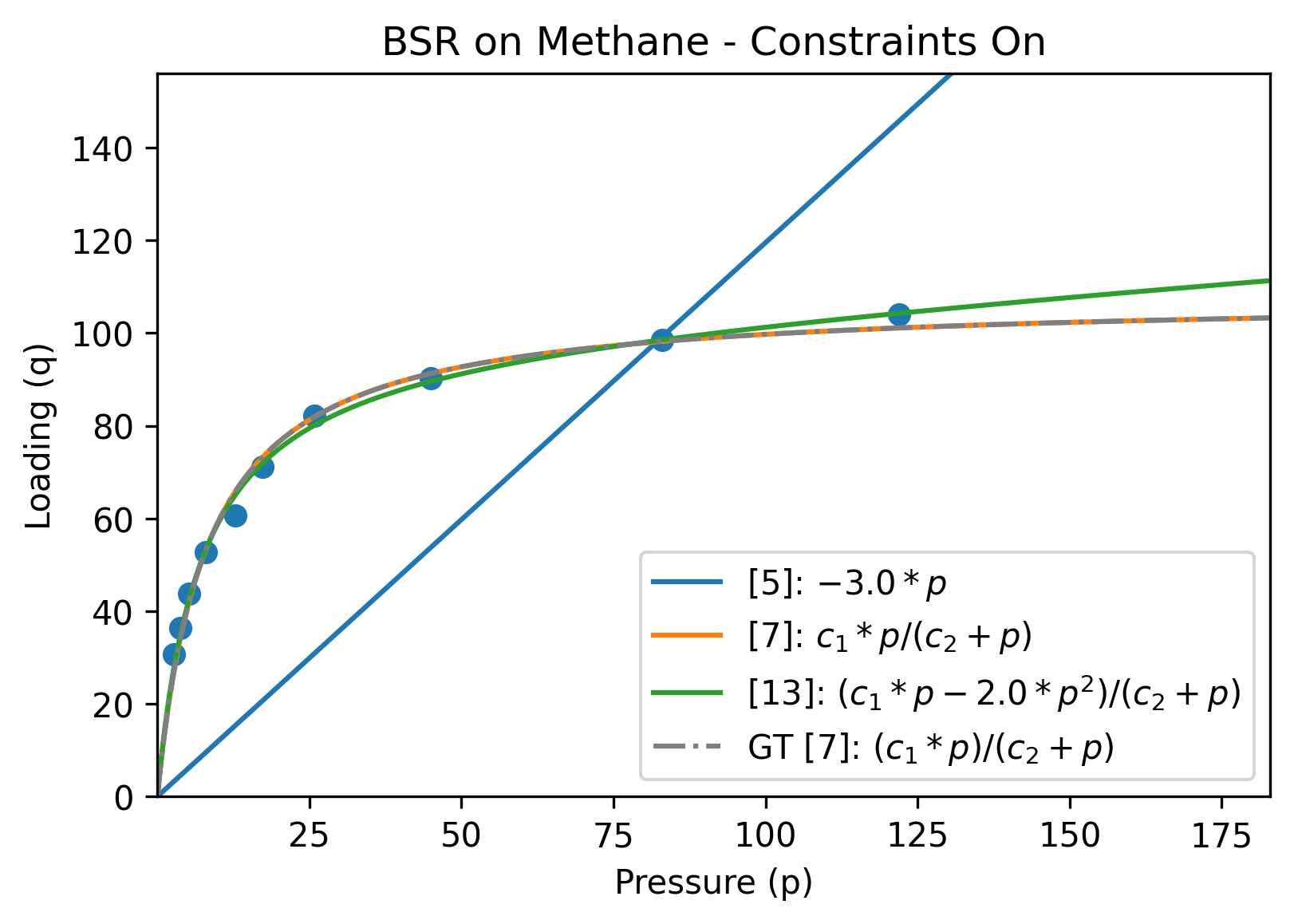}
        \label{fig:BSRMethaneOnIsotherm}
    \end{subfigure}%
    \vspace{-2\baselineskip}
    \begin{subfigure}[b]{0.45\textwidth}
        \caption{}
        \includegraphics[trim={0 0 0 0.6cm}, width=\textwidth]{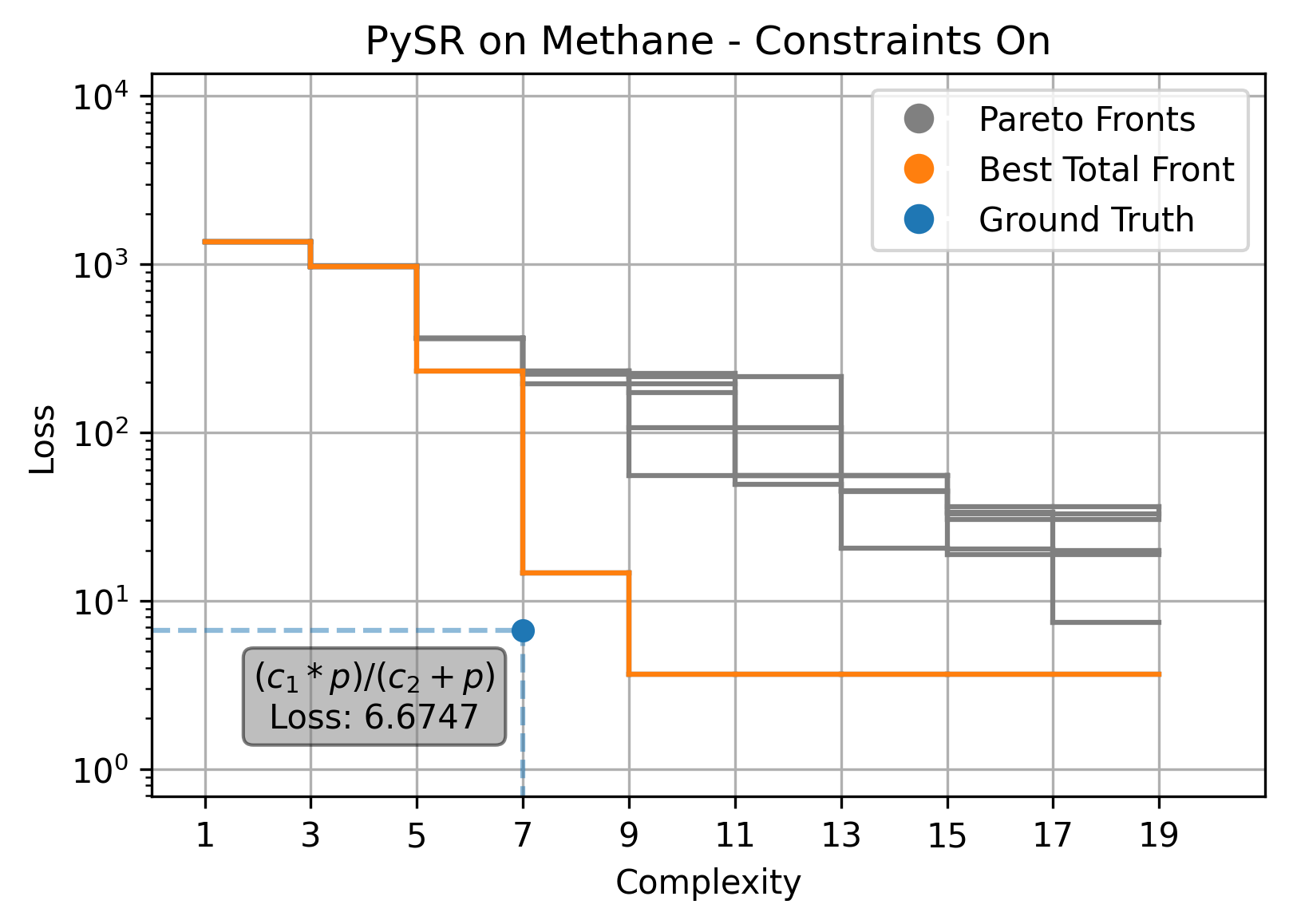}
        \label{fig:PySRMethaneOnPareto}
    \end{subfigure}%
    \hfill
    \begin{subfigure}[b]{0.45\textwidth}
        \caption{}
        \includegraphics[trim={0 0 0 0.6cm}, width=\textwidth]{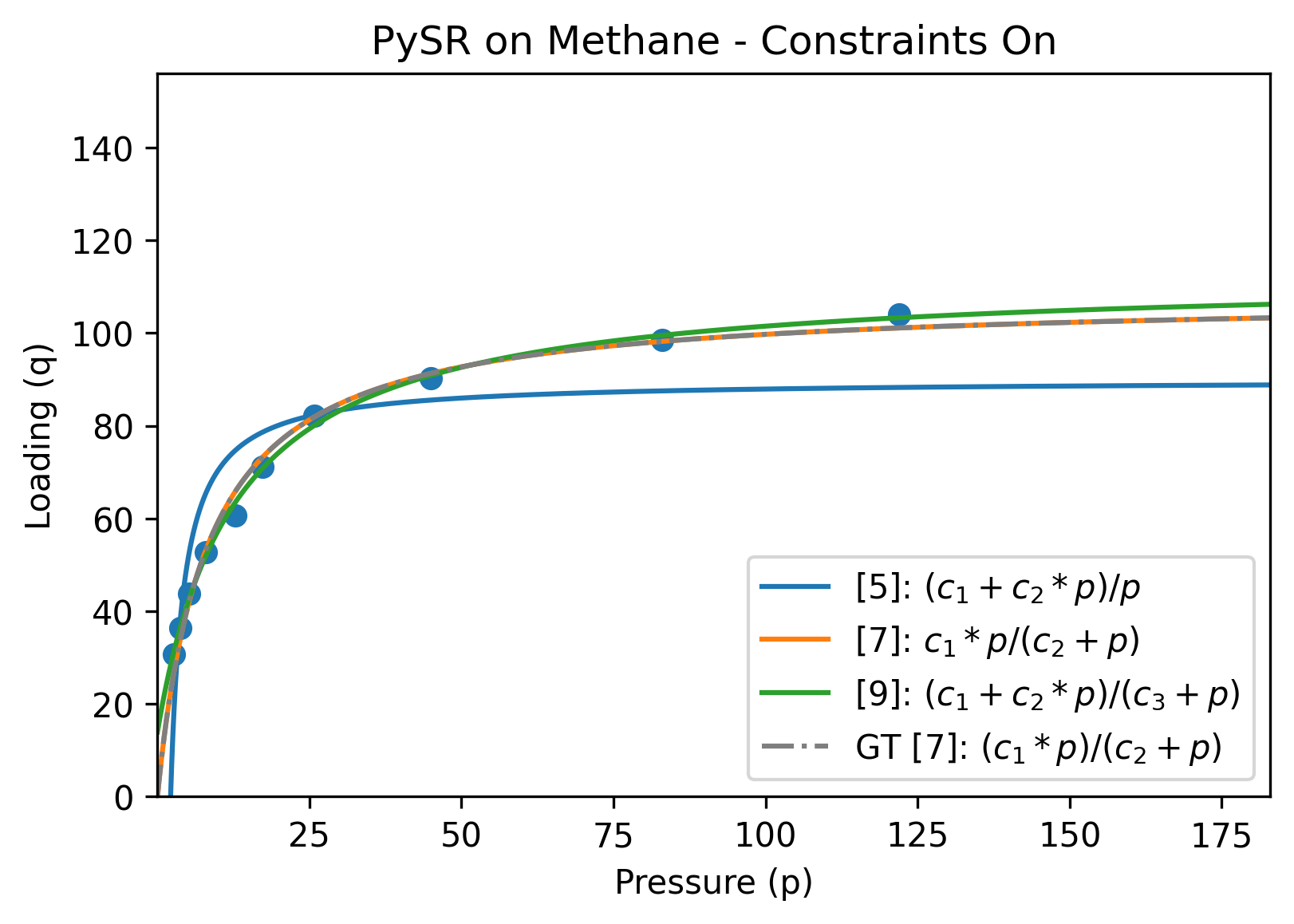}
        \label{fig:PySRMethaneOnIsotherm}
    \end{subfigure}%
    \vspace{-1\baselineskip}
    \caption{BSR and PySR on the Methane dataset.  The left column shows combined Pareto fronts across 8 runs and the right column shows interesting isotherms found at the defining corners of those Pareto fronts.  The constraints are disabled in the top four subplots and enabled in the bottom four.  The rows alternate between BSR and PySR.}
    \label{fig:Methane}
\end{figure}

While the results are mostly similar for the methane dataset, there are some important differences.  Like with the nitrogen dataset, BSR finds the ground truth without constraints enabled while PySR does not.  This is apparent in the Pareto fronts (Fig. \ref{fig:BSRMethaneOffPareto} and Fig. \ref{fig:PySRMethaneOffPareto}). In this case, PySR finds an expression with complexity 9 with more accuracy than the ground truth, though with an extra constant in the numerator, it violates the thermodynamic constraints (Fig. \ref{fig:PySRMethaneOffIsotherm}). Imposing the constraints penalized the loss for this expression relative to the ground truth, but not enough to overcome the increased accuracy (Fig. \ref{fig:PySRMethaneOnIsotherm}). As with nitrogen, BSR does a better job of finding expressions that pass the constraints, even when they are not enabled, as it finds 33\% passing the first and 51\% passing the second (where PySR finds 4.1\% and 48\% respectively).

\subsection{Isobutane Dataset}
\begin{figure}[H]
    \centering
    \begin{subfigure}[b]{0.45\textwidth}
        \caption{}
        \includegraphics[trim={0 0 0 0.6cm}, width=\textwidth]{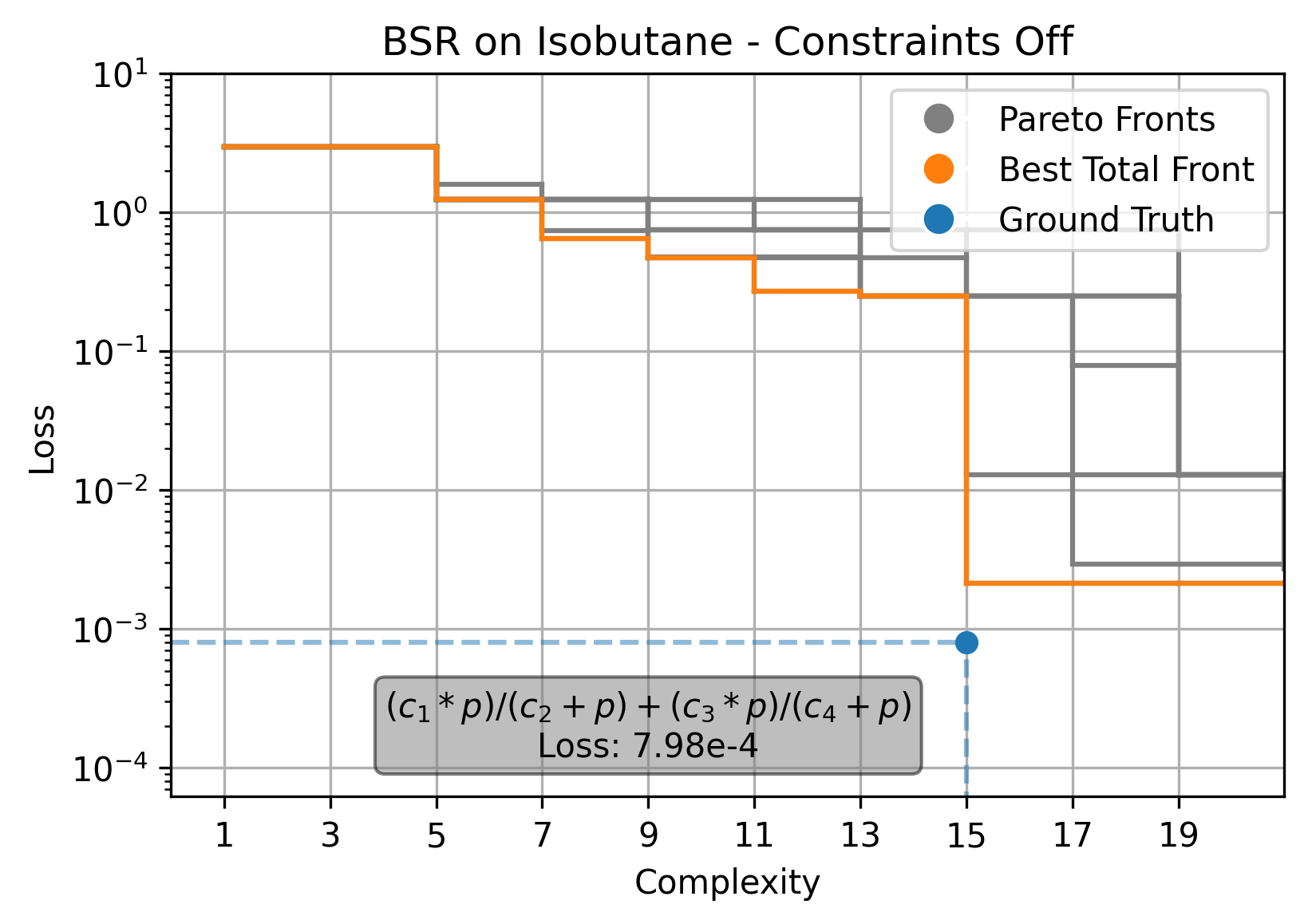}
        \label{fig:BSRIsobutaneOffPareto}
    \end{subfigure}%
    \hfill
    \begin{subfigure}[b]{0.45\textwidth}
        \caption{}
        \includegraphics[trim={0 0 0 0.6cm}, width=\textwidth]{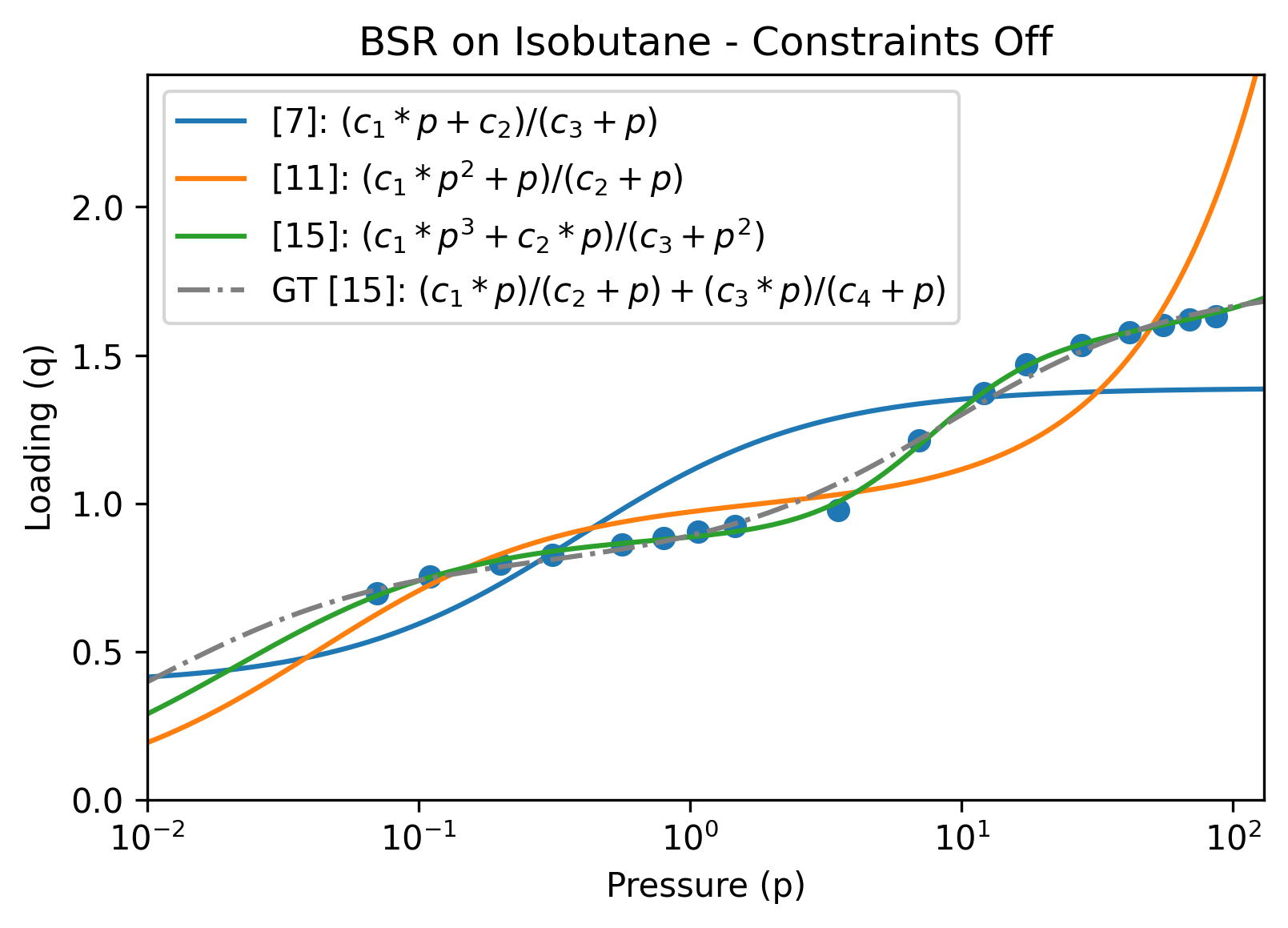}
        \label{fig:BSRIsobutaneOffIsotherm}
    \end{subfigure}%
    \vspace{-2\baselineskip}
    \begin{subfigure}[b]{0.45\textwidth}
        \caption{}
        \includegraphics[trim={0 0 0 0.6cm}, width=\textwidth]{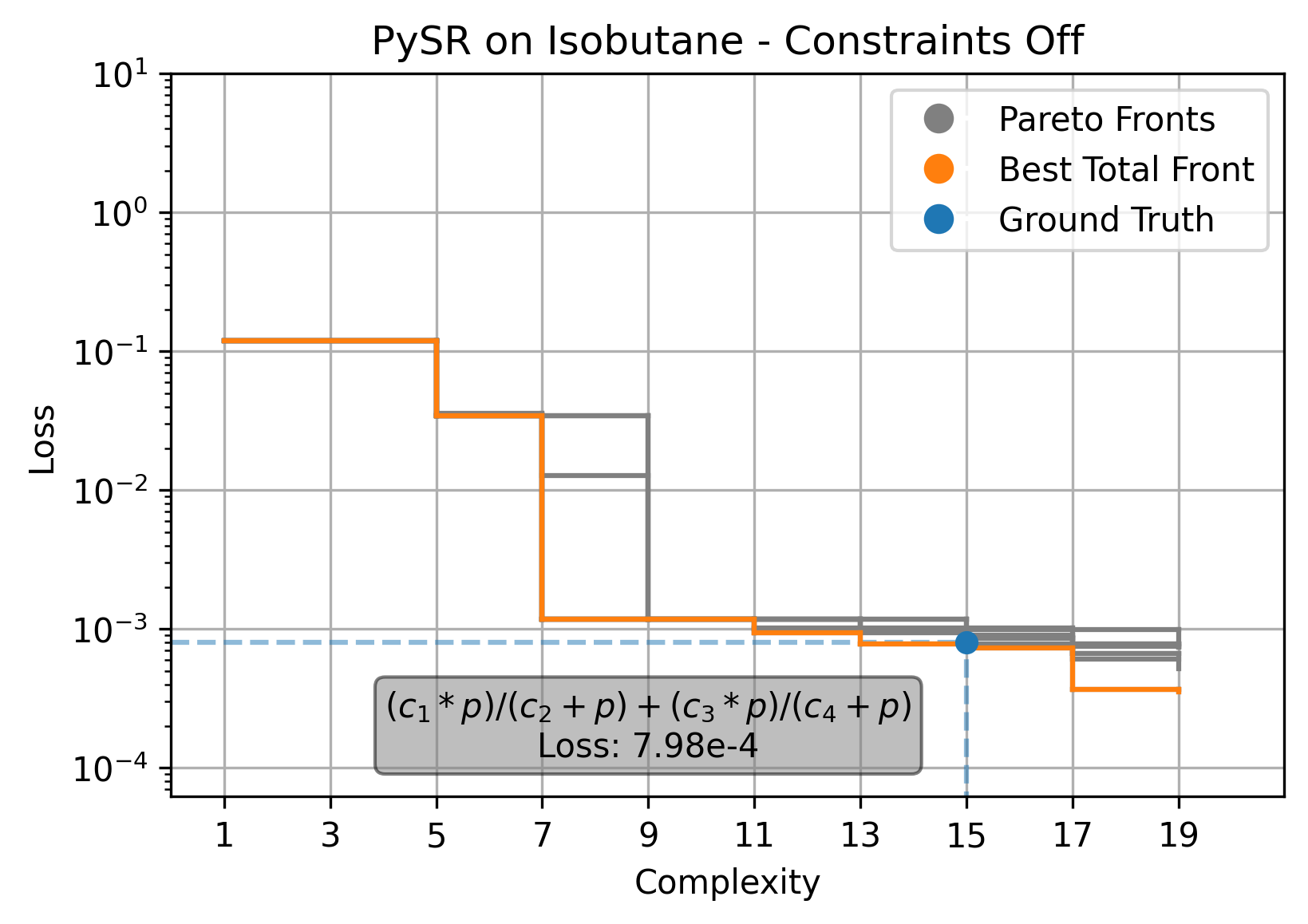}
        \label{fig:PySRIsobutaneOffPareto}
    \end{subfigure}%
    \hfill
    \begin{subfigure}[b]{0.45\textwidth}
        \caption{}
        \includegraphics[trim={0 0 0 0.6cm}, width=\textwidth]{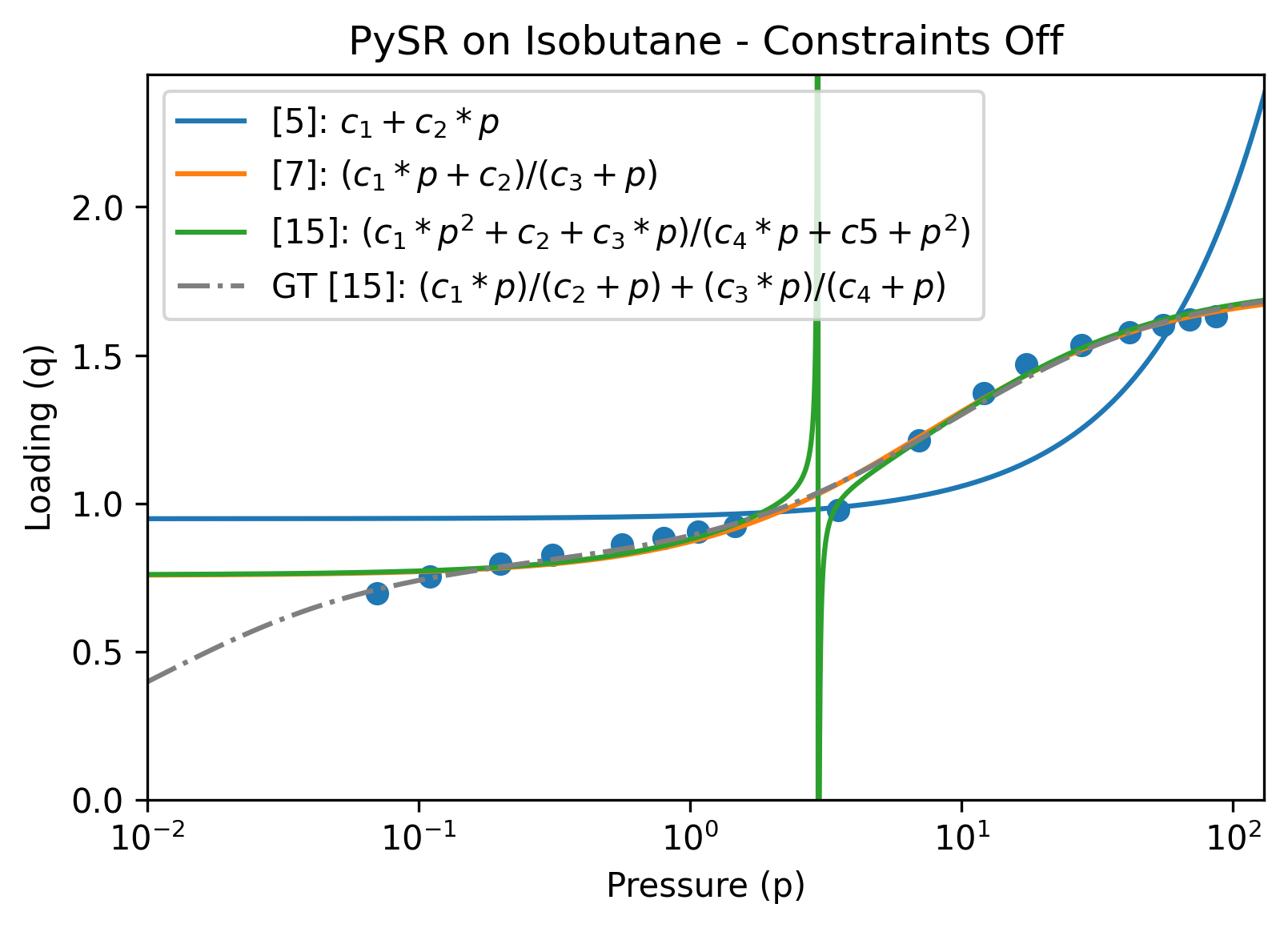}
        \label{fig:PySRIsobutaneOffIsotherm}
    \end{subfigure}%
    \vspace{-2\baselineskip}
    \begin{subfigure}[b]{0.45\textwidth}
        \caption{}
        \includegraphics[trim={0 0 0 0.6cm}, width=\textwidth]{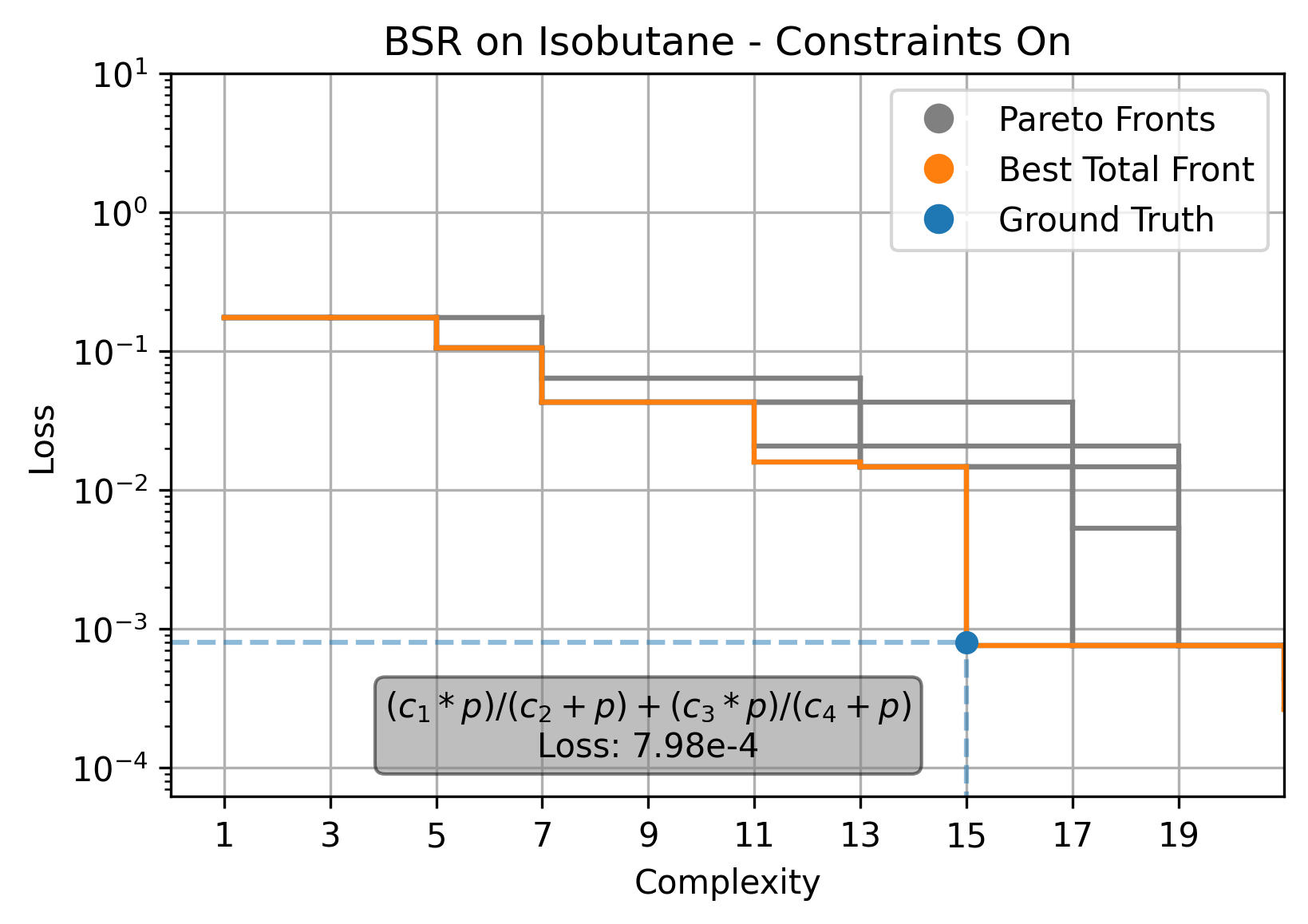}
        \label{fig:BSRIsobutaneOnPareto}
    \end{subfigure}%
    \hfill
    \begin{subfigure}[b]{0.45\textwidth}
        \caption{}
        \includegraphics[trim={0 0 0 0.6cm}, width=\textwidth]{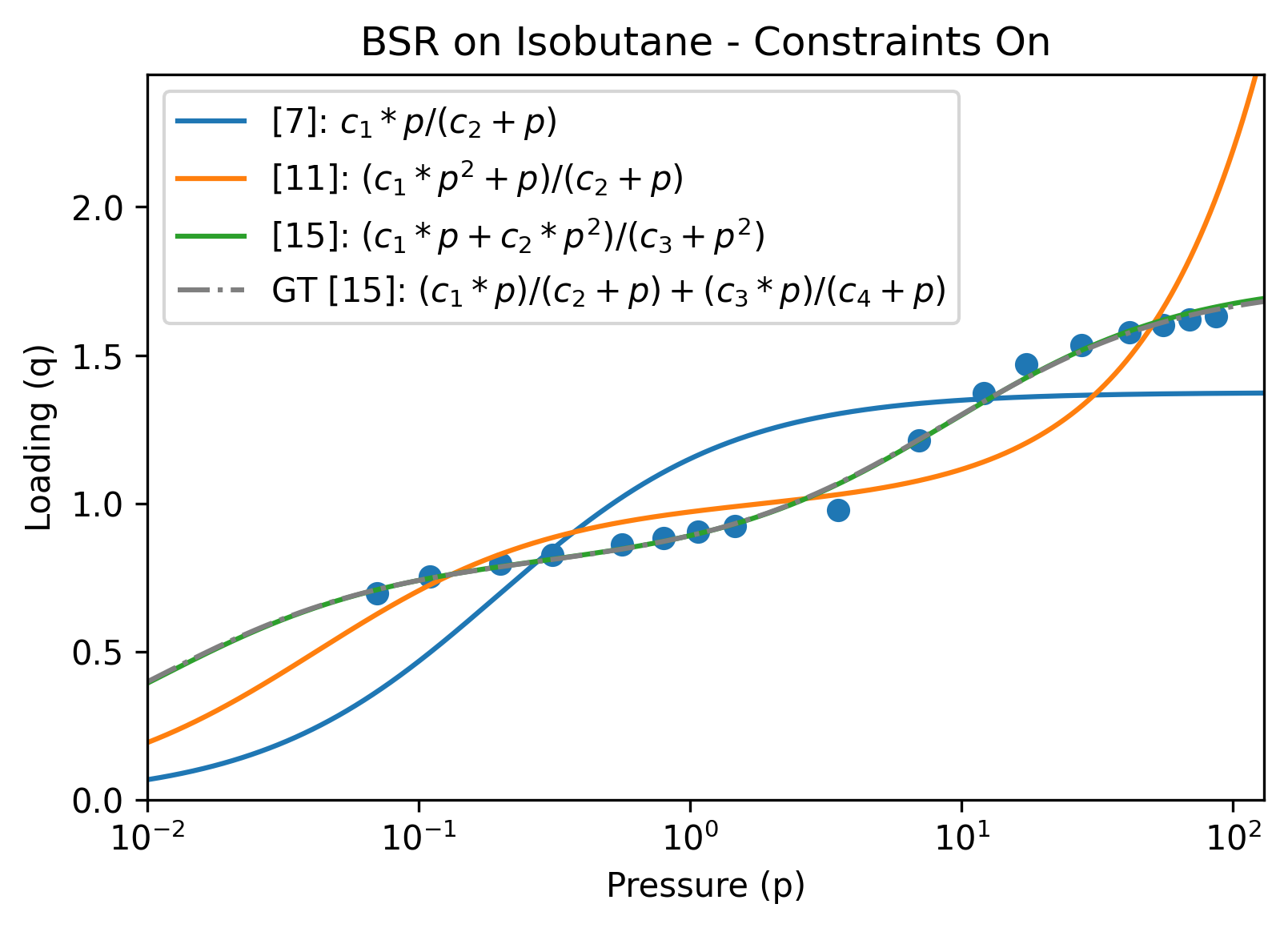}
        \label{fig:BSRIsobutaneOnIsotherm}
    \end{subfigure}%
    \vspace{-2\baselineskip}
    \begin{subfigure}[b]{0.45\textwidth}
        \caption{}
        \includegraphics[trim={0 0 0 0.6cm}, width=\textwidth]{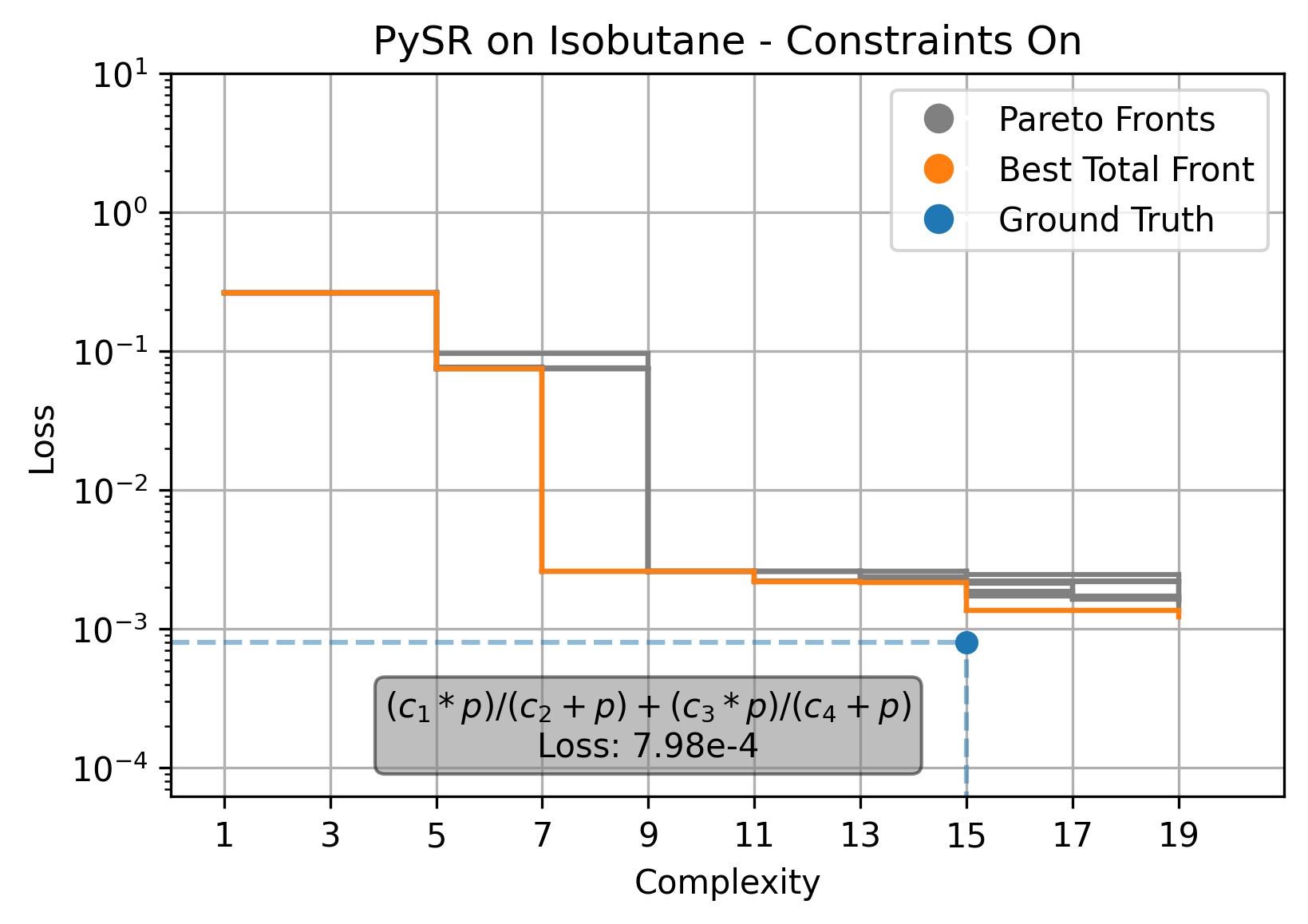}
        \label{fig:PySRIsobutaneOnPareto}
    \end{subfigure}%
    \hfill
    \begin{subfigure}[b]{0.45\textwidth}
        \caption{}
        \includegraphics[trim={0 0 0 0.6cm}, width=\textwidth]{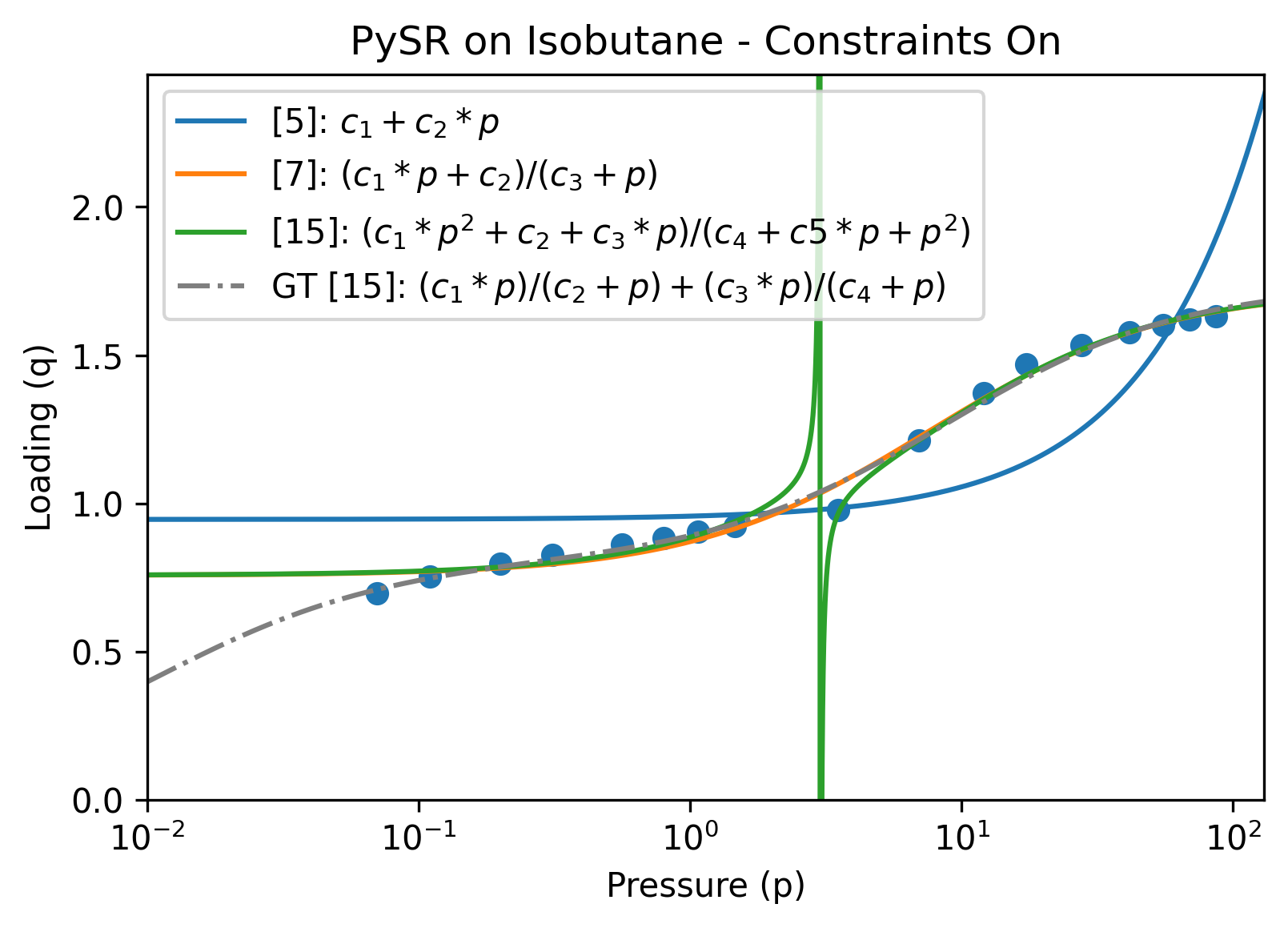}
        \label{fig:PySRIsobutaneOnIsotherm}
    \end{subfigure}%
    \vspace{-1\baselineskip}
    \caption{BSR and PySR on the Isobutane dataset.  The left column shows combined Pareto fronts across 8 runs and the right column shows interesting isotherms found at the defining corners of those Pareto fronts.  The constraints are disabled in the top four subplots and enabled in the bottom four.  The rows alternate between BSR and PySR.}
    \label{fig:Isobutane}
\end{figure}

Unlike the methane and nitrogen datasets (Fig. \ref{fig:Nitrogen} and \ref{fig:Methane}) which are best modeled by the Langmuir isotherm, the isobutane dataset (Fig. \ref{fig:Isobutane}) is best modeled by the dual-site Langmuir isotherm, which has twice the complexity. Despite this significant complexity, the dual-site Langmuir isotherm is not significantly more accurate than many expressions shorter than it. This is best seen in Fig. \ref{fig:PySRIsobutaneOffPareto} and \ref{fig:PySRIsobutaneOnPareto} which show the Pareto fronts for PySR with constraints off and on respectively. In both plots, expressions with half the complexity reach almost the same accuracy, creating a plateau from complexity 7 onward. This is also shown well in the corresponding isotherm plots which show that the expressions found at complexity 7 match the data as well as the ground truth. Importantly, these expressions do not satisfy the thermodynamic constraints.

Unlike PySR, BSR does not find expressions with accuracy close to the ground truth until the same complexity. 

For BSR, including constraints shifts the whole Pareto front down (Fig. \ref{fig:BSRIsobutaneOffPareto} to Fig. \ref{fig:BSRIsobutaneOnPareto}), indicating that more accurate expressions were found at many complexity levels. While PySR did not find accurate expressions consistent with the constraints, BSR did. 
In this case, BSR finds the ground truth expression while PySR does not. This is not apparent on either the Pareto fronts or isotherm plots however, because the accuracy of the expression found is about 10x worse than the fit ground truth and the best expressions found at that complexity.  This is likely because, while the ground truth is found, the form it was originally produced in (before being simplified) is much more complex.

In PySR, penalizing expressions that violate constraints actually led to populations of equations that violated constraints two and three more often, with a decrease of about 10\% in each case (see Table ~\ref{tab:PercentPassing}). This was surprising -- we anticipated that imposing penalties would lead to fewer violating expressions, but the opposite occurred. For BSR as well, including constraints in the prior actually led to a decrease in expressions satisfying the second constraint (from 46\% to 36\%), and a slight increase in the first and third constraints. 


\subsection{BET Dataset}
\begin{figure}[H]
    \centering
    \begin{subfigure}[b]{0.45\textwidth}
        \caption{}
        \includegraphics[trim={0 0 0 0.6cm}, width=\textwidth]{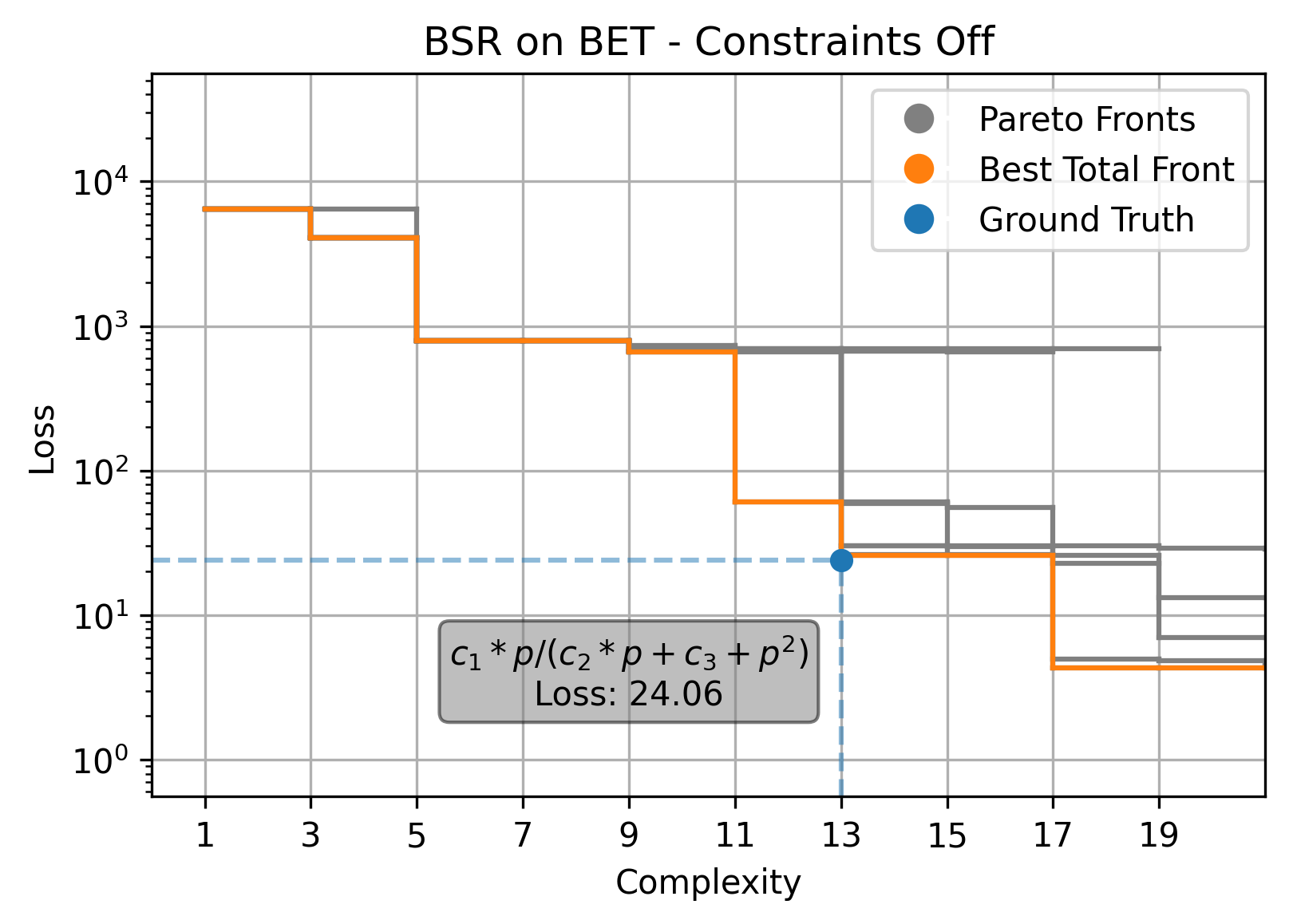}
        \label{fig:BSRBETOffPareto}
    \end{subfigure}%
    \hfill
    \begin{subfigure}[b]{0.45\textwidth}
        \caption{}
        \includegraphics[trim={0 0 0 0.6cm}, width=\textwidth]{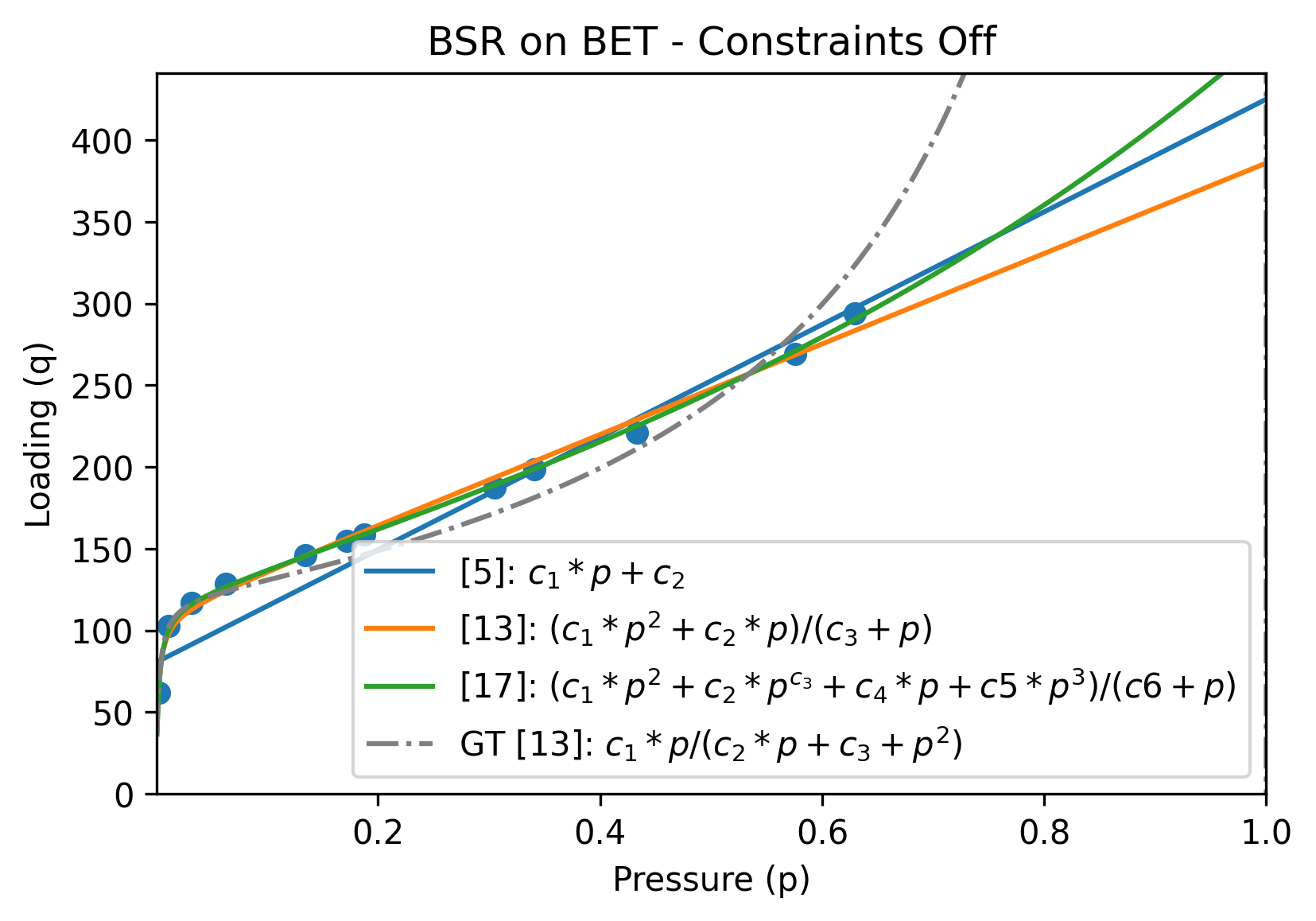}
        \label{fig:BSRBETOffIsotherm}
    \end{subfigure}%
    \vspace{-2\baselineskip}
    \begin{subfigure}[b]{0.45\textwidth}
        \caption{}
        \includegraphics[trim={0 0 0 0.6cm}, width=\textwidth]{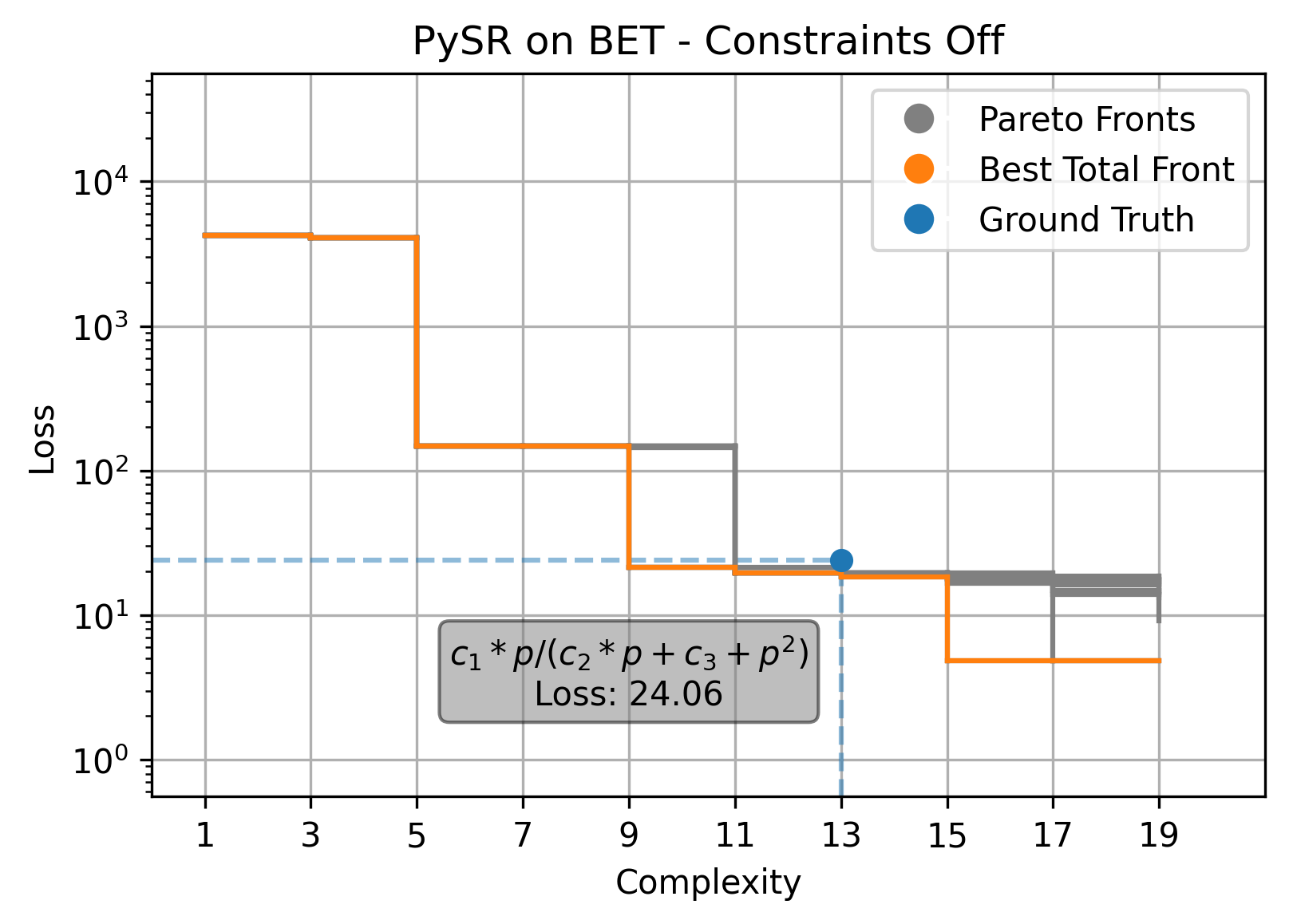}
        \label{fig:PySRBETOffPareto}
    \end{subfigure}%
    \hfill
    \begin{subfigure}[b]{0.45\textwidth}
        \caption{}
        \includegraphics[trim={0 0 0 0.6cm}, width=\textwidth]{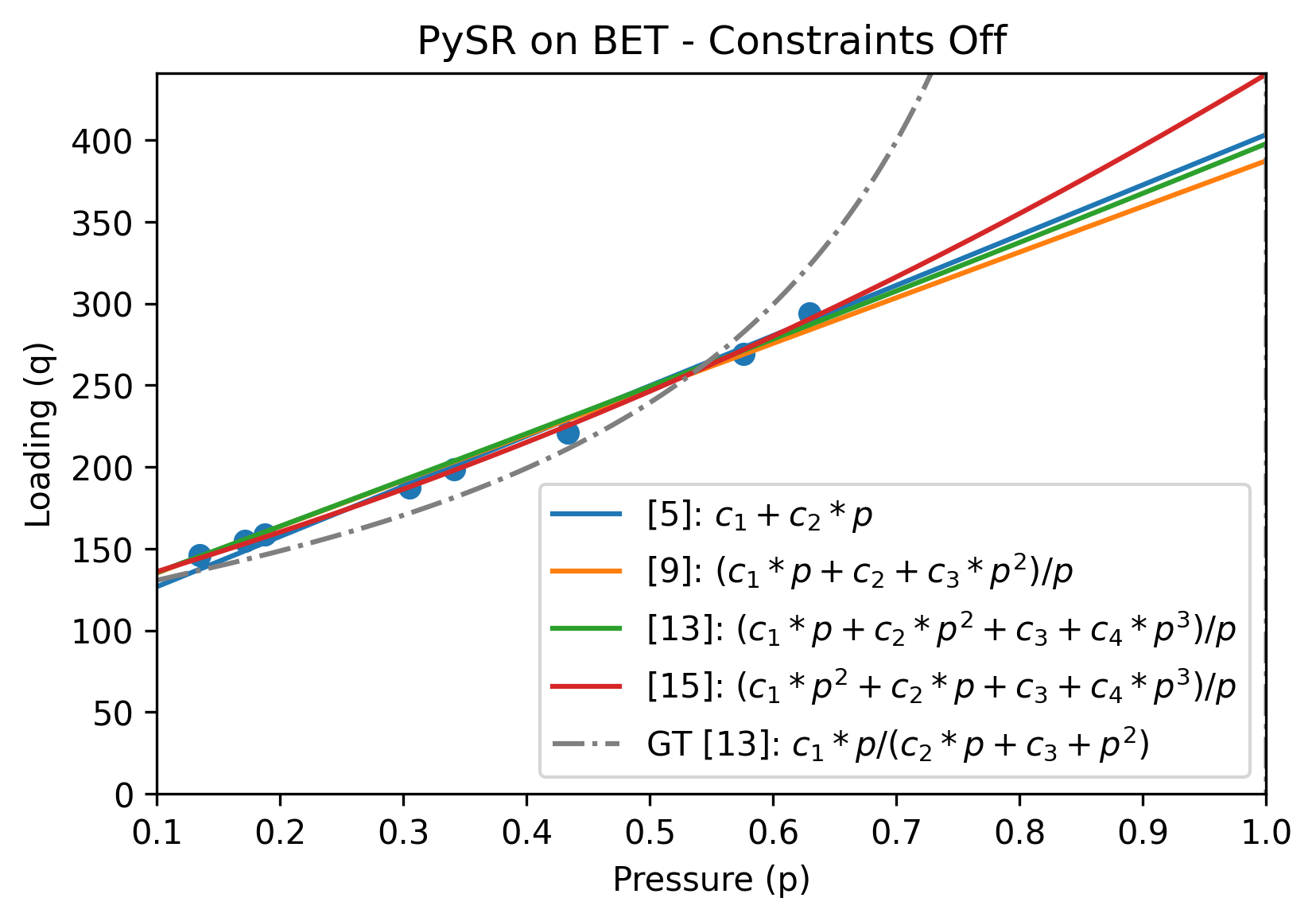}
        \label{fig:PySRBETOffIsotherm}
    \end{subfigure}%
    \vspace{-2\baselineskip}
    \begin{subfigure}[b]{0.45\textwidth}
        \caption{}
        \includegraphics[trim={0 0 0 0.6cm}, width=\textwidth]{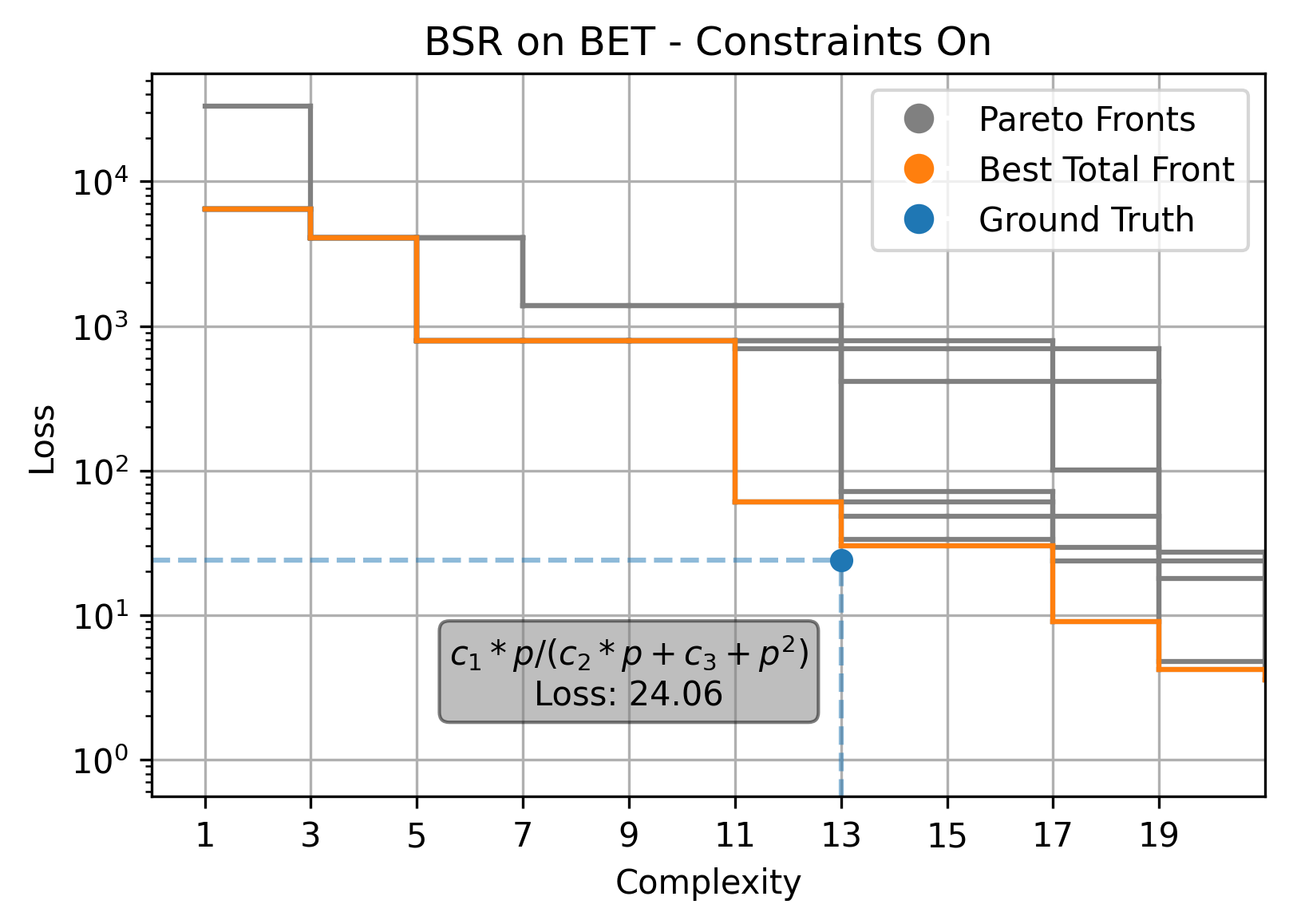}
        \label{fig:BSRBETOnPareto}
    \end{subfigure}%
    \hfill
    \begin{subfigure}[b]{0.45\textwidth}
        \caption{}
        \includegraphics[trim={0 0 0 0.6cm}, width=\textwidth]{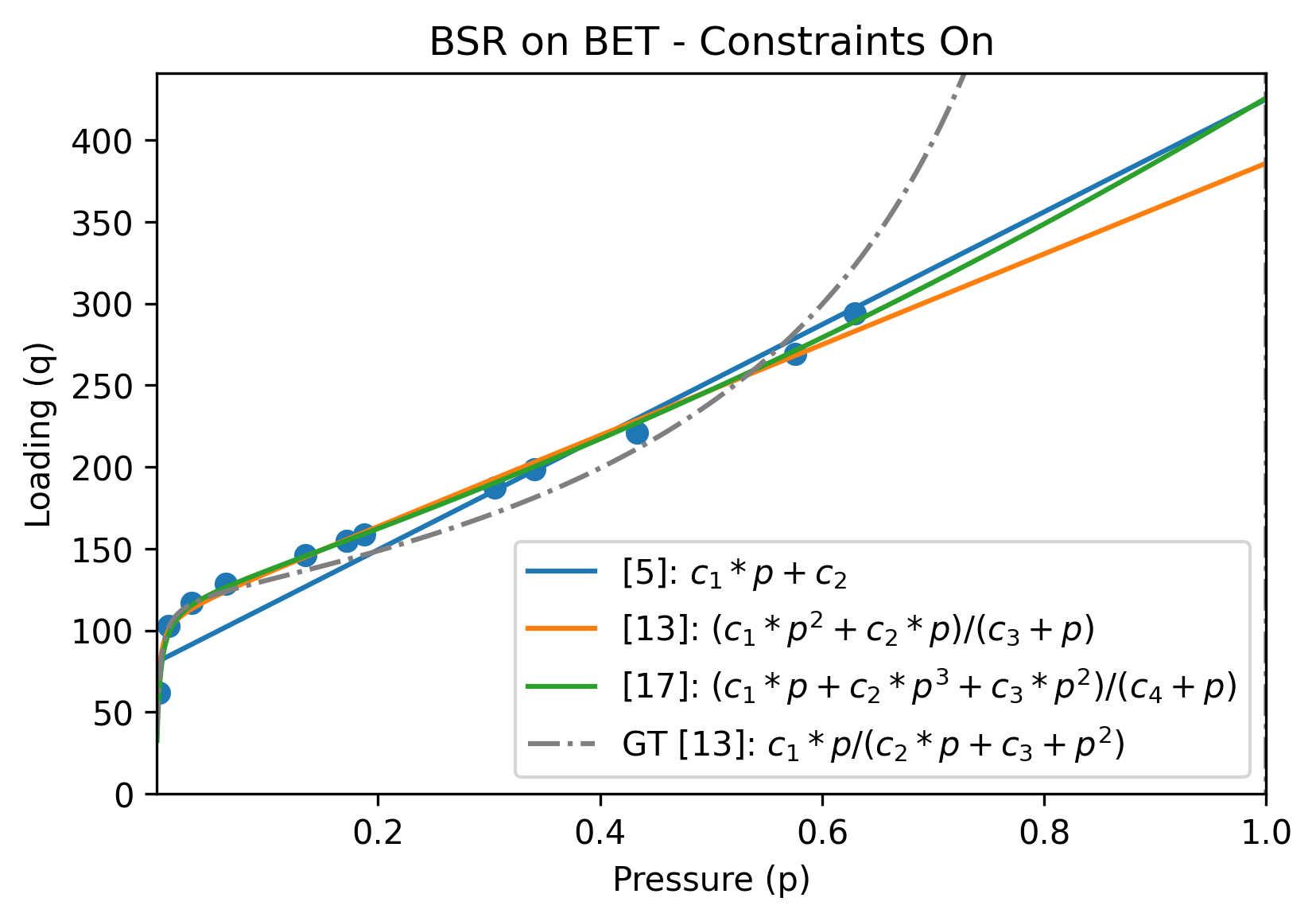}
        \label{fig:BSRBETOnIsotherm}
    \end{subfigure}%
    \vspace{-2\baselineskip}
    \begin{subfigure}[b]{0.45\textwidth}
        \caption{}
        \includegraphics[trim={0 0 0 0.6cm}, width=\textwidth]{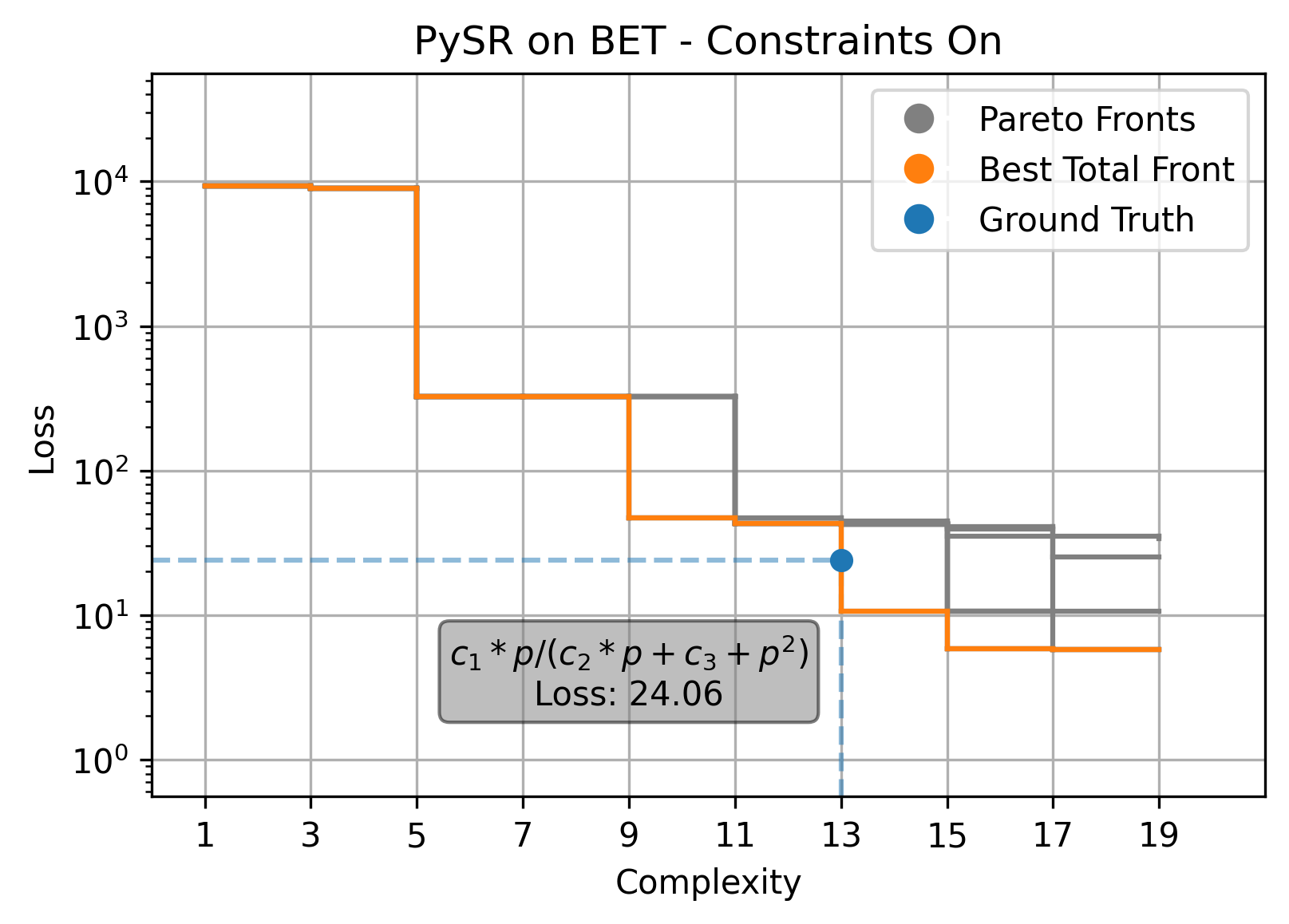}
        \label{fig:PySRBETOnPareto}
    \end{subfigure}%
    \hfill
    \begin{subfigure}[b]{0.45\textwidth}
        \caption{}
        \includegraphics[trim={0 0 0 0.6cm}, width=\textwidth]{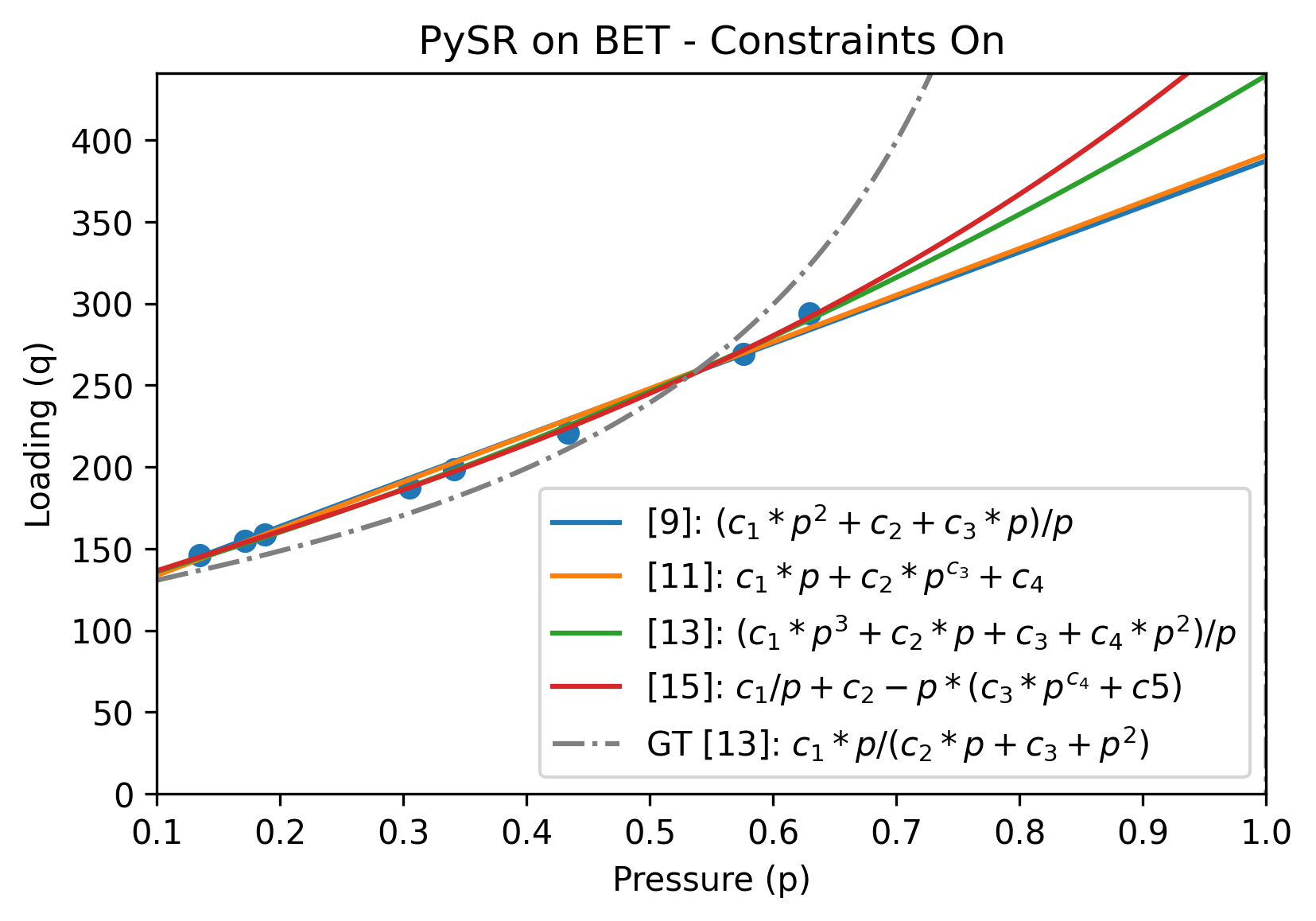}
        \label{fig:PySRBETOnIsotherm}
    \end{subfigure}%
    \vspace{-1\baselineskip}
    \caption{BSR and PySR on the BET dataset.  The left column shows combined Pareto fronts across 8 runs and the right column shows interesting isotherms found at the defining corners of those Pareto fronts.  The constraints are disabled in the top four subplots and enabled in the bottom four.  The rows alternate between BSR and PySR.}
    \label{fig:BET}
\end{figure}

The BET dataset is unique because the ground truth expression diverges to infinity as the pressure approaches 1 (pressure in this case is relative vapor pressure, $p/p^{\mathrm{sat}}$; the vapor being adsorbed becomes a liquid as $p/p^{\mathrm{sat}} \rightarrow 1$. So in this case, the third constraint (that it is monotonically non-decreasing) no longer holds for all pressure (seen in Fig. \ref{fig:BET}).
Nonetheless, we found that whether or not constraints were enabled, many of the most accurate expressions generated by PySR for this dataset pass the third constraint (78.65\% without constraints and 81.29\% with), contrary to the ground truth theory.
Furthermore, PySR satisfies the first two constraints less frequently with constraints on compared to with constraints off. One possible explanation for this behavior is that the dataset itself is more easily fit by expressions with expressions that are monotonically non-decreasing, at least from the perspective of the PySR algorithm. Overall, while PySR can find accurate expressions for the BET dataset, it fails to find expressions that also follow the constraints, even when they are enabled.

In contrast, BSR did not generate many expressions that were monotonically nondecreasing, and the incorporation of constraints had a substantial effect on the search. Specifically, the second constraint is passed about 92\% of the time both with it enabled and disabled and the portion passing the first constraint increases dramatically from 16\% to 85\% once it is enabled. This leads to a large number of models which agree with the requisite constraints for BET, but none of these are the ground truth rediscovered. Instead, many expressions with close to (or better than) the accuracy of the ground truth are found by both algorithms in both cases, none of the isotherms plotted appear similar. The asymptote at a partial pressure of 1 is not replicated by any similarly accurate expressions and the slight curve of the ground truth in the middle of the dataset is also absent. These results together seem to indicate that the constraints, while thermodynamically correct, do not provide enough information (or even provide contradictory information) for rediscovering the BET ground truth expression.


\begin{table}[!h]
\renewcommand{\arraystretch}{2}
\resizebox{\textwidth}{!}{%
\begin{tabular}{|c|c|c|c|c|c|c|c|}
\hline
Dataset   & Constraints Active & BSR C1  & BSR C2  & BSR C3  & PySR C1 & PySR C2 & PySR C3 \\ \hline
Nitrogen  & False              & 37\% & 67\% & 0.46\%  & 2.2\%  & 33\% & 46\% \\ \hline
Nitrogen  & True               & 72\% & 73\% & 19\% & 10\% & 41\% & 59\% \\ \hline
Methane   & False              & 33\% & 51\% & 0.51\%  & 4.1\%  & 48\% & 61\% \\ \hline
Methane   & True               & 59\% & 59\% & 2.5\%  & 5.7\%  & 54\% & 62\% \\ \hline
Isobutane & False              & 24\% & 46\% & 0.45\%  & 3.4\%  & 65\% & 68\% \\ \hline
Isobutane & True               & 36\% & 36\% & 1.3\%  & 5.5\%  & 56\% & 58\% \\ \hline
BET       & False              & 16\% & 92\% & 0.09\%  & 6.2\%  & 35\% & 79\% \\ \hline
BET       & True               & 85\% & 92\% & 2.4\%  & 4.7\%  & 30\% & 81\% \\ \hline
\end{tabular}%
}
\caption{Percentage of expressions generated passing each of the three constraints.  Results are shown across both SR methods, all datasets and with constraints active and disabled.}
\label{tab:PercentPassing}
\end{table}


\section{Discussion}

\subsection{Effectiveness}

This work highlights that sometimes, a stochastic search through equation space finds equations that are superior to ground truth expressions -- in our case achieving comparable accuracy to the ground truth expressions while also being less complex (shorter expression length).  This is particularly observed in the case of isobutane (Fig. \ref{fig:Isobutane}); both with constraints on and off, PySR finds the expression $\frac{c_1p + c_2}{c_3 + p}$, which fits the data well but diverges from the ground truth as it approaches 0.  But many of these expressions, while consistent with the data, are inconsistent with thermodynamics, as they violate our tested constraints.  We have demonstrated that accounting for these constraints in the search process can guide the population or distribution of expressions toward thermodynamically consistent expressions, and aid in the identification of the ground truth expression. 
Sometimes this leads to more consistent equations, 
and sometimes it doesn't improve the search at all. 

\subsection{Exceptions To Constraints}


While the three constraints presented in this work do follow from a broader thermodynamic theory, not all isotherm models in this work satisfy all the constraints.  Specifically, the BET isotherm does not satisfy the third constraint because it reaches an asymptote as $p/p_0$ approaches 1.  While this does break the monotonicity constraint, it also seems reasonable when considering what $p_0$ represents (the pressure at which the adsorbate becomes a liquid and the interaction fundamentally changes).  This raises the question: what constraints to include and why?
The decision to test if expressions pass the third constraint necessarily excludes the BET model because, while it is theoretically grounded, it only applies from in the range $(0, p_0)$. Furthermore, the data used does not extend past that range so expressions that do pass the third constraint may not appear much different in terms of what is relevant. We recommend carefully examining what constraints one may want to test and in what places they should actually be checked. Introducing incorrect constraints may hinder the search with our biases, and prevent algorithms from discovering phenomena outside our assumptions.

\subsection{Computational Complexity / Runtime}

\begin{figure}[ht]
    \centering
    \includegraphics[width=\textwidth]{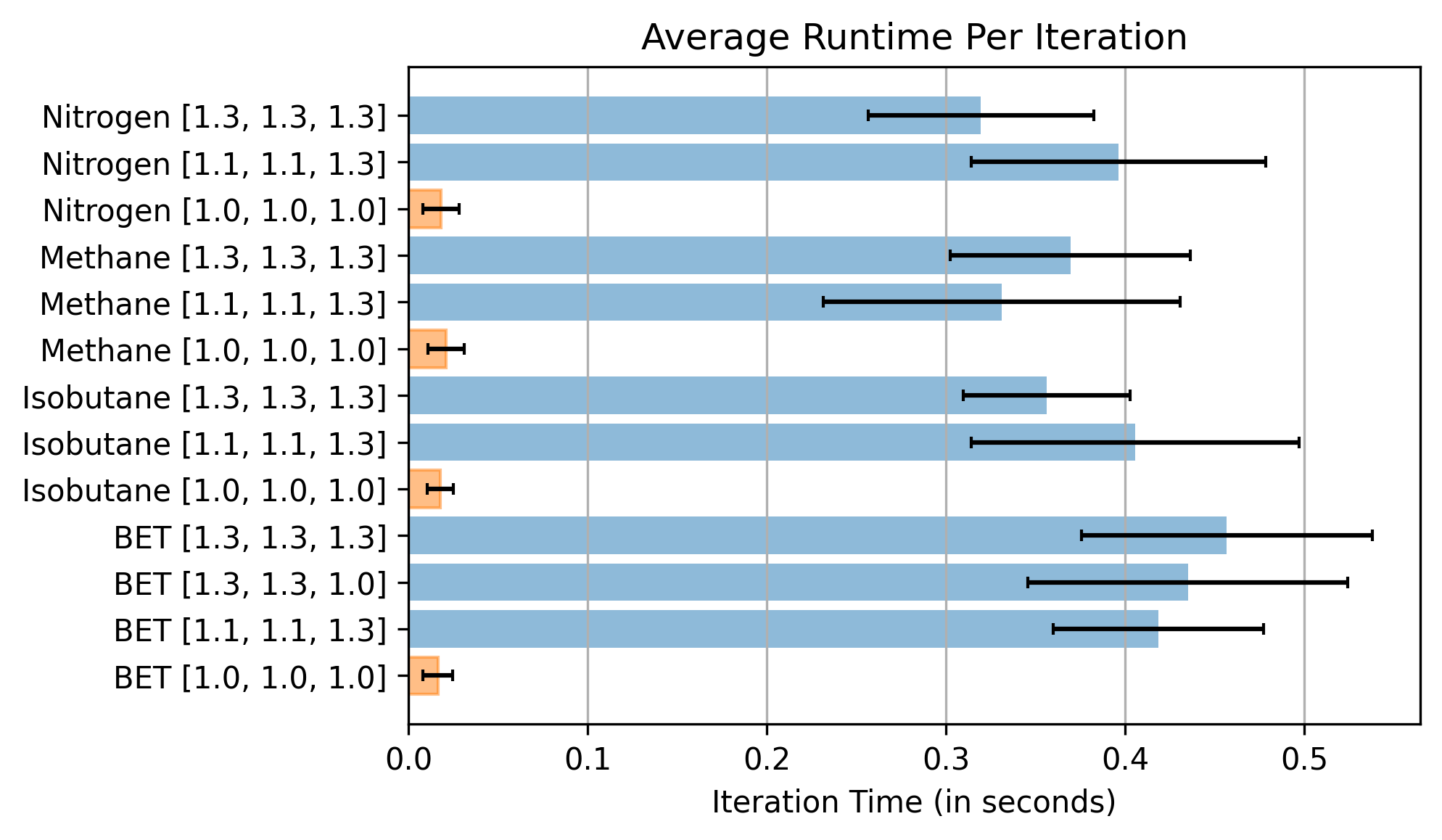}
    \caption{Average runtimes across all datasets and combinations of thermodynamic constraint penalties.  Runs with all penalties set to 1.0 are highlighted in orange.  Standard deviation is shown by error bars at the top of each bar.}
    \label{fig:runtimes}
\end{figure}

As seen in Fig. \ref{fig:runtimes}), consideration of constraints increases runtime by an order of magnitude; this is even after we carefully integrated the computer algebra system into the SR algorithms to reduce overhead, and leveraged memory to avoid redundant checks on expressions previously visited. On average, numerically checking models is much faster than manipulating them symbolically (especially for larger expressions) -- checking \emph{every} new expression is quite expensive. This is unfortunately necessary if the constraints are to be considered as an integral part of the search. If a cheaper solution is needed, the search can be performed without constraints, and constraints checked after the fact. In fact, this approach enables even more elaborate methods of considering background knowledge, such as comparing against complex, multi-premise background theories using an automated theorem prover \cite{cornelio_ai_2021}.

\subsection{Challenges around complexity, simplification, and canonical form}

\begin{figure}[ht]
    \centering
    \includegraphics{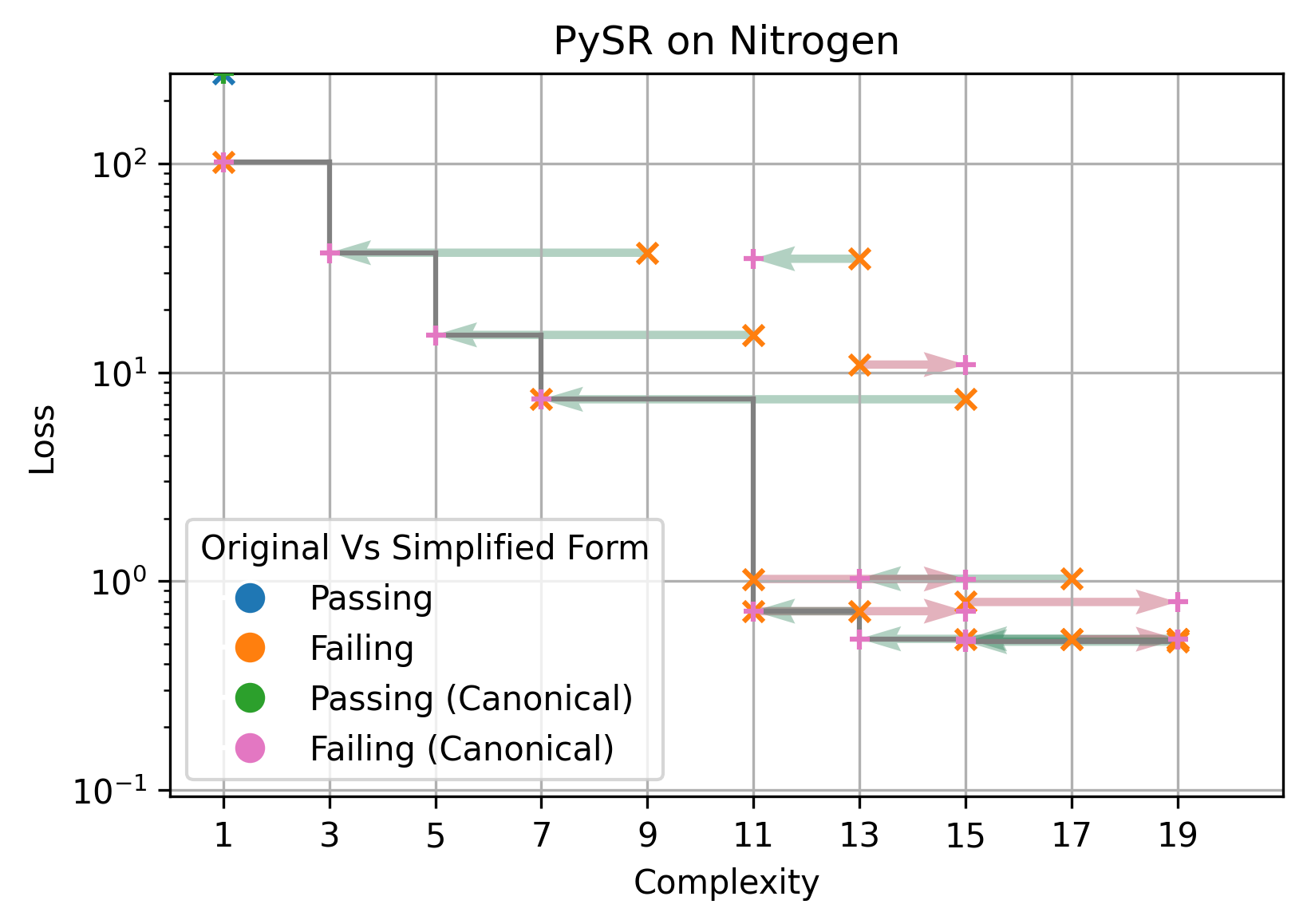}
    \caption{The effect of the canonical form checking function on a Pareto front showing the results from PySR on the Nitrogen dataset.  Points on this plot are marked as ``Passing" only if they pass all constraints checked.}
    \label{fig:PySRSimplification}
\end{figure}

In this work, simplification is necessary in order to identify whether a generated expression matches the ground truth and to assign generated expressions an appropriate complexity. We augmented SymPy’s “simplify” function, to shorten the numerous rational expressions we generated into a "canonical form" (details in the Supporting Information).
While some methods such as BSR attempt simplification during runtime, PySR does not because of the added computation needed per expression. Generating a "canonical form" for expressions generated by PySR sometimes increases, and sometimes decreases, the complexity. Some expressions are generated as complex expression trees that are much more complex than their canonical forms  (Fig.~\ref{fig:PySRSimplification}).

Simplification is a crucial challenge of this work because complexity plays a significant role in SR.  After all, models are compared via accuracy and complexity to make decisions during the search. A single model may very well take on different scores / likelihoods because of how it is written, influencing not just its standing, but subsequent steps in the search. Ideally, every model would always be written in the simplest form, but this is computationally intractable in some circumstances \cite{richardson_identity_1994}.  Because of this, comparing functions based on behavior (symbolic constraints) may be more appropriate, because limiting behavior is invariant to the numerous ways an expression can be written.

\subsection{Reducing Underspecification Through Inductive Biases}

Machine learning researchers at Google recently highlighted the role of underspecification in machine learning pipelines \cite{damour_underspecification_2020}.
They suggested that one way to combat underspecification is to use credible inductive biases to allow for selection from otherwise similarly effective models, and that these constraints should have little negative effect on accuracy if selected correctly.
In this work, we find expressions that are roughly equivalent in terms of accuracy and complexity but have different functional forms, leading to different behavior outside the range of the data -- signatures similar to those discussed in \cite{damour_underspecification_2020}.
We find that adding thermodynamic constraints can help improve the search for good expressions, but this doesn't necessarily restrict the hypothesis space in the same way that inductive biases do; we were unable to effectively search with hard constraints, and so our hypothesis space still included expressions that are inconsistent with constraints.
Instead, we can reduce the hypothesis space after the search is complete; by rejecting accurate-but-inconsistent expressions using our background knowledge, we improve on the issues of underspecification.
Nonetheless, for datasets with reasonably complex behavior, there still exist multiple distinct thermodynamically-consistent expressions of similar accuracy and complexity. The space of equations defined by the limited number of operators considered here, even for one dimensional datasets, is just that vast! 




\section{Conclusions}

In this work, we couple a computer algebra system to two symbolic regression algorithms in order to check the consistency of generated expressions with background knowledge. 
We find that including appropriate mathematical constraints can improve search effectiveness or break the search entirely, depending on the dataset and implementation details. Although computational costs increase by an order of magnitude, tightly integrating SR with a computer algebra system is a practical way to check for constraints on each expression generated during the search.

We have shown that consideration of constraints helps in rediscovering ground-truth isotherm models from experimental data, including the Langmuir and the dual-site Langmuir isotherms (though the dual-site Langmuir isotherm was not identified on the Pareto front, it was present in the generated models). In contrast, the BET isotherm was not rediscovered; more accurate and concise models were generated instead, and the most meaningful model (BET) was consequently missed. 
We found that Bayesian Symbolic Regression is a more effective and intuitive platform for incorporating symbolic constraints in a Bayesian prior, rather than by modifying the fitness function in traditional genetic algorithms; the resulting populations of expressions were more attuned to the constraints with BSR. 
Finally, though background knowledge can screen out accurate yet inconsistent solutions, symbolic regression pipelines remain underspecified in our context, capable of generating multiple distinct solutions with similar performance and adherence to constraints.

\section{Acknowledgements}

We thank Marta Sales-Pardo and Roger Guimer\`{a} for discussions about the Bayesian Machine Scientist, and Miles Cranmer for assistance with PySR. This material is based upon work supported by the National Science Foundation under Grant No. \#2138938, as well as startup funds from the University of Maryland, Baltimore County.

\section{Supporting Information}

The modified version of PySR and the code used to run it are both available on GitHub.  The PySR code was forked from the original repository on June 6th, 2020 and is available at \href{https://github.com/CharFox1/SymbolicRegression.jl}{https://github.com/CharFox1/SymbolicRegression.jl}. The code for running PySR, parsing its output, and plotting the results, and is available at \href{https://github.com/ATOMSLab/pySR_adsorption}{https://github.com/ATOMSLab/pySR\_adsorption}.
The modified version of BMS code used in this paper is available at \href{https://github.com/ATOMSLab/BayesianSymbolicRegression}{https://github.com/ATOMSLab/BayesianSymbolicRegression}.

The Supporting Information includes 1) further description of the adsorption models considered here, 2) further discussion of the changes we implemented in PySR and BMS codes to implement thermodynamic constraint checking, including pseudocode for new algorithms, 3) description of our pipeline for collecting and analyzing generated expressions, 4) further discussion of the nuances around identifying of ``interesting'' expressions in automated pipelines and algorithms for simplification and pattern-matching, 5) details of constant fitting, 6) experiments comparing runtime of algorithms on different datasets, and 7) details of the testing environment on the UMBC supercomputer.

\newpage
\bibliographystyle{unsrt}
\bibliography{SRPaperCitations}

\newpage

\section{Supporting Information}

\subsection{Langmuir}
The Langmuir isotherm model was originally presented in 1918 and remains in common use today \cite{langmuir_adsorption_1918}.

\begin{gather}
    q = \frac{q_mK_Lp}{1+K_Lp} \quad \text{ or } \quad \theta = \frac{K_Lp}{1+K_Lp}
\end{gather}

The original Langmuir isotherm relates the volume adsorbed onto the surface ($q_e$), the adsorption strength ($K_L$), the gas pressure $p$ and the maximum adsorption capacity ($q_m$) \cite{wang_adsorption_2020} \cite{langmuir_adsorption_1918}.  The isotherm is often more simply written with only the fractional adsorption $\theta_A$, for the fraction of the surface that is occupied by adsorbed molecules. In practice, $q_m$ is not known \emph{a priori}, so we aim to rediscover the expression for $q_e$. However, this expression is \emph{never} found by SR, because an equivalent expression with shorter length (7 vs. 11) can be written as
\begin{gather}
    q = \frac{q_mK_LP}{1+K_LP} \rightarrow q = \frac{c1 * p} {(c2 + p)}
\end{gather}

The basic premise of the Langmuir model is to represent the adsorbent (the surface atoms or molecules are adsorbing onto) as a simple surface with some number of free and full spots where the adsorbate could stick.  The model is not concerned with rough or porous surfaces, molecules that may take up multiple ``sites" or otherwise interact with each other after being adsorbed or, most importantly, any stacking of molecules.  
Wang et. al. classify the Langmuir model as a chemical adsorption model because it is only concerned with mono-layer adsorption in which a chemical bond is formed between the adsorbent and adsorbate \cite{wang_adsorption_2020}.

\subsection{Dual-Site}
The Langmuir model can be extended to describe more complex materials. Specifically, if a surface has two types of adsorption sites (with different properties),  another term can be added to obtain the ``dual-site Langmuir" model:

\begin{gather}
    q = \frac{q_aK_Ap} {(1 + K_Ap)} + \frac{q_bK_Bp} {(1 + K_Bp)} \rightarrow q = \frac{c1 * p} {(c2 + p)} + \frac{c3 * p} {(c4 + p)}
\end{gather}


In this case, because there are two unique Langmuir terms, there are also two unique terms for both the maximum volume that can be adsorbed ($q_a$ and $q_b$) as well as the ``Langmuir Constants" ($K_A$ and $K_B$). This model is considered to be the ground truth for the isobutane dataset because the adsorbent material, the MFI zeolite, has two distinct adsorption sites for isobutane \cite{vlugt_molecular_1999}.


\subsection{BET} \label{sec:BET}
The BET model extends the Langmuir model to consider multi-layer adsorption \cite{brunauer_adsorption_1938}. Beyond the first layer, van der Waals forces attract adsorbates toward the surface as well as to adsorbed molecules. Because of this, Wang et. al. classifies the BET isotherm as a physical model as opposed to a chemical one \cite{wang_adsorption_2020}. The main BET model assumes infinite possible layers; alternative forms can be derived for a finite maximum number of layers \cite{brunauer_adsorption_1938}. With a max of one layer, it simplifies to the Langmuir model (which only models one layer), with $n$ layers, it becomes significantly more complex (and unlikely to be found via SR). Fortunately the form from $n = \infty$ is concise and can be simplified to:


\begin{gather}
    q = \frac{v_m*c*(p/p_0)}{(1-p/p_0)*(1+ (c-1)p/p_0)}
\end{gather}

with $p_0$ being the saturated vapor pressure, and $v_m$ and $c$ being constants describing monolayer adsorption and interaction energies in the system. Data can be provided to SR as $p$ or as $p/p_0$; using the second choice is a form of \emph{a priori} feature selection, which assumes we know that pressure should be normalized to $p_0$. Because we don't assume this, $p_0$ becomes a third constant fit by SR. SR generally fits this to an incorrect value, since it is unconstrained to match $p_0$ to anything physical, and prioritizes fit to the given data.

\subsection{Thermodynamic Constraint Functions}
Three constraint checking functions (which follow from the three thermodynamic constraints introduced previously in section \ref{sec:OriginalThermoConstraints}) were developed using the Python library SymPy \cite{10.7717/peerj-cs.103}.  Each function returns a Boolean TRUE or FALSE, depending on if its constraint is met or not. While these functions are useful for examining the expressions generated after a run, simply discarding an expression for failing one or more constraint during a run can severely hinder search potential by cutting off intermediary steps between better expressions that may also pass the constraints. Because of this, each function has a corresponding weight that allows it to act as a ``soft constraint". Specifically, a constraint will not affect the score of an expression if it is passed but will multiply (worsen) that score if it is not.  This approach (as implemented in PySR) is detailed in Algorithm \ref{alg:constraints}. Algorithm \ref{alg:monotonic} describes our test whether our function is monotonically non-decreasing, using SymPy.



\begin{algorithm}[h]
\caption{Modified genetic algorithm}\label{alg:constraints}
\begin{algorithmic}
\Require{$tree, dataset, options$}
\Ensure{$score, loss$}
\Function{scoreFunc}{$tree, dataset, options$}
\State $loss \gets \text{evalLoss}(tree, dataset, options)$ \Comment{Traditional error calculation}
\State $penalties \gets options.penalties$
\If{$penalties \text{ not empty }$} \Comment{Skip slow calculation if possible}
\State $expr, var \gets \text{parseTree}(tree)$
\For{$ p \in penalties \text{ and } tf \in thermoFunctions$}
\If{$tf(expr, var) = \text{False }$}
\State $loss \gets loss * p$
\EndIf
\EndFor
\EndIf
\State $score \gets \text{scoreFunc}(loss, tree, options)$ \Comment{Factor in complexity}
\State \Return $score, loss$
\EndFunction
\end{algorithmic}
\end{algorithm}

\begin{algorithm}[H]
\caption{Monotonic Non-decreasing Check}\label{alg:monotonic}
\begin{algorithmic}
\Require{$expr, var, start, stop$} 
\Ensure{$passing$ (If the expression is monotonically non-decreasing)}
\Function{montonicNondecreasing}{$expr, var, start, stop$}
\If{$expr \text{ is constant}$}
\State \Return True
\EndIf
\State $turningPoints \gets \text{inflection points (zeros) of } expr$ \Comment{Calculated using SymPy}
\If{$turningPoints \text{is empty}$}
\If{$expr(stop) - expr(start) \geq 0$} \Comment{Measure slope}
    \State \Return True
\Else
    \State \Return False
\EndIf
\EndIf
\State $turningPoints \gets [start, turningPoints, stop]$ \Comment{Include bounds}
\For{$\text{each sequential pair of points } p, q \in turningPoints$}
\If{$expr(q) - expr(p) \leq 0$} \Comment{Measure slope between zeros}
    \State \Return False
\EndIf
\EndFor
\State \Return True
\EndFunction
\end{algorithmic}
\end{algorithm}

\subsection{Implementation of Thermodynamic Constraints} 

In order to maintain parity between PySR and BMS, the Julia library PyCall was used.  This allowed the same Python code that checked thermodynamic constraints in BMS to be used within the Julia code of the SymbolicRegression.jl library (the back-end of PySR).  Beyond the concern of a fair comparison across platforms, PyCall showed itself to be somewhat necessary for this work.  As of writing, there is no Julia library close to as full-featured as SymPy (and SymPy can have some serious limitations/difficulties).  One method explored was using the Julia library MATLAB.jl to call MATLAB code from within Julia in a similar way to how PyCall works with Python code. While this eventually worked (MATLAB does have a robust symbolic math toolbox), this proved very difficult to set up due to intricacies in how the two languages store data. MATLAB was not significantly faster than Python in our tests.

One downside to the current structure of our version of SR.jl and the languages it relies on is that it must be run in distributed mode instead of threaded mode (meaning multiple processes must be created, requiring more overhead).  This is due to the fact that PyCall expects only one Julia instance to attempt to use Python at once.  With multiple processes, a new instance of the PyCall Julia package is created for each process, thus avoiding memory access conflicts.

\subsection{PySR Data Collection}
Though PySR generates and handles many expressions, accessing complete information about all members and all populations over time proved difficult. PySR's goal each iteration is to produce a Pareto front showing what it has found to be the most accurate expression at each complexity level. This means the vast majority of expressions in a population are not shown (although some may be duplicates). Indeed, if an expression never makes it to the Pareto front at any point during the run, by the definition of the search, it is not as important. However, this prevented us from investigating whether expressions congruent with the ground truth were generated and discarded, because they did not land on the Pareto front (due to insufficient parameter optimization or simplification). 

Nonetheless, to track the expressions we could easily access (generally hundreds of Pareto fronts per run), the full output text is parsed using Python and stored in a Pandas DataFrame \cite{reback_pandas-devpandas_2022}. The values collected from the raw text are the expression itself, score, loss, complexity, runtime, iteration and run number. While this is a significant amount of data already, there are a few interesting attributes yet to be calculated -- specifically those relating to the simplified form of the expressions and those related to the thermodynamic constraints. To make further parsing manageable, once the simplified form of each expression is generated, only the most accurate expression with that form is retained. This step often reduces the number of rows in the DataFrame by about 1000x. The simplified form of each expression is calculated both with original (optimized/fit) and substituted constants. The new complexity of the simplified form may or may not be smaller than the original complexity so both attributes are kept.  Finally, the thermodynamic constraints that may have been used to guide the search are checked for each expression.

\subsection{Identifying ``interesting" expressions}

An important component of this work is identification of interesting or meaningful expressions generated by the SR algorithms. Expressions that satisfy constraints are more interesting than those that do not, but the most meangingful expressions are those which are \emph{derivable} \cite{cornelio_ai_2021}. In our case, this is indicated by being able to be simplified to the canonical form of an already known expression such as the Langmuir or BET adsorption isotherms. Unfortunately, determining if one expression is equivalent to another is a very difficult, and sometimes undecidable problem. In fact, while not applicable to this work due to our limits on operators and constants used, Richardson's theorem shows that, including certain operators and the transcendental numbers $\pi$ and $e$, it may be impossible to show that one expression is equivalent to another \cite{richardson_identity_1994}. Consequently, we identify exact expressions by either automatically finding a direct match to the canonical form, or by manually inspecting expressions that satisfy all relevant constraints and given low errors, but this does not guarantee that we find all instances of rediscovered ground truth expressions.


\subsection{Simplification Function}
We were first inclined to leave the constants as symbolic and treat them as variables (e.g. c1, c2...), but we found that  SymPy did not reliably simplify fully symbolic expressions of even modest size. We consequently substituted either the fitted parameters from the search, or by replacing symbolic constants with a sequence of prime numbers. We simplify the expression first using SymPy's default tool, but this doesn't simplify some expressions to their shortest form. For example, if the resulting expression is rational (has only integer exponents and no division by zero), it can be written as a fraction and possibly simplified further. Specifically, if there is a common factor between the numerator and denominator because they are both degree 1 or larger, it may be possible to remove a leading constant, providing a simpler expression than that from SymPy's default tool. For example,
\begin{gather}
    \frac{2x}{3x^2 + 4x + 5} \rightarrow \frac{x}{(3/2)x^2 + 2x + 5/2} \quad \text{or} \quad \frac{(2/3)x}{x^2 + (4/3)x + 5/3}
\end{gather}

\begin{gather}
    \frac{c_1x}{c_2x^2 + c_3x + c_4} \rightarrow \frac{x}{c_1x^2 + c_2x + c_3} \quad \text{or} \quad \frac{c_1 x}{x^2 + c_2x + c_3}
\end{gather}
In this case, an expression of 4 constants is reduced to an expression of 3 constants.

Some expressions have the same constant appearing in multiple places. For example, the BET expression from literature, as well as its simplified form, have constants $p_0$ or $c_2$ appearing multiple times:
\begin{gather}
    \frac{v_m*c*(p/p_0)}{(1-p/p_0)*(1+ (c-1)p/p_0)} \rightarrow \frac{c_1*p}{(p^2 + c_2*p + c_2)}
\end{gather}
In our case, we accept solutions with the same form as the ground truth, but with non-unique constants. The expression $c1*p / (p^2 + c_2*p + c_3)$ is equal in complexity (using our metrics) while have more ability to fit the data; if we required $c_2 = c_3$, we would never obtain the ground truth.

\begin{singlespace}
\begin{algorithm}[H]
\caption{Simplification Function}\label{alg:cap}
\begin{algorithmic}
\Require{$expr, vars, pars$} 
\Ensure{$can$ (Canonical form of expression)}
\Function{simplify}{$expr, vars, pars$}
\State $expr \gets \text{expression}$ 
\State $vars \gets \text{variables}$ 
\State $pars \gets \text{parameters}$ 
\For{$p \textbf{ in } pars$}
\State $p \gets \text{prime number}$ \Comment{Substitute constants with unique primes}
\EndFor
\State $\text{Simplify using SymPy}$
\If{$expr \text{ is rational}$}
\State $num, denom \gets expr \text{ as fraction}$ \Comment{Calculated using SymPy}
\If{$\text{degree}(num) > \text{degree}(denom)$}
\State $factor \gets \text{ leading term of } num$
\Else
\State $factor \gets \text{ leading term of } denom$
\EndIf
\State $expr \gets \frac{num/factor}{denom/factor} $ \Comment{Remove common factor}
\EndIf
\State \Return $expr$
\EndFunction

\end{algorithmic}
\end{algorithm}
\end{singlespace}

\subsection{Fitting Constants in PySR}

By default, PySR uses Nelder-Mead optimization to fit constants for expressions as they are generated \cite{cranmer_milescranmerpysr_2021} \cite{mogensen_julianlsolversoptimjl_2022}.  Nelder-Mead is well suited to optimizing parameters for arbitrary generated expressions because it does not require any derivative or gradient information. The algorithm evaluates the function at $n+1$ points where $n$ is the number of parameters. The next point is selected by finding the point with the highest function evaluation and looking to the opposite side of the remaining points (a simplex). This iteratively moves the worst point to a likely better location, eventually moving towards a local optimum regardless of how the function is evaluated.  

One downside to this method is that it is not guaranteed (or even expected) to find global minima.  This issue is usually rectified by allowing for multiple starts from random locations, and selecting the best result.  In PySR, each expression gets 8 attempts by default, randomizing the parameters between each.  This is typically enough but there are occasionally cases where an expression should be much more accurate than it is due to poorly fitted constants.  

To fit ground-truth expressions and check constants in post-processing, we also applied Nelder-Mead optimization using SciPy \cite{virtanen_scipy_2020}. We allowed many more iterations in post-processing to ensure ground truths and interesting functional forms were most optimized.

\subsection{Runtime}
An important consideration when examining the effectiveness of the thermodynamic constraints in guiding SR is the impact on computation time. While not extreme, symbolic math (in SymPy) can be slow, especially compared to the otherwise efficient and optimized Julia code running behind the PySR front-end. The following plot shows the difference in iteration time (the time between one Pareto front and the next being printed by PySR) across different datasets, and more importantly, across different constraint penalties. It shows that runtime without SymPy is fast and not dependent on the dataset, and that it is increased by about an order of magnitude when checking constraints.

\subsection{Environment} 


Testing was done on both the batch and cpu2021 partitions of the UMBC High Performance Computing Facility (https://hpcf.umbc.edu/).  This allowed for the longer runtimes and larger parameter exploration necessitated by adding thermodynamic constraints with variable penalties. 

While BMS is entirely based in Python, SymbolicRegression.jl (the backend for PySR) is entirely in Julia.  Because of the need to use SymPy, the Julia package PyCall.jl was used to allow Python code and libraries to be run by Julia (at the time this work was completed, symbolic math libraries in Julia could not evaluate the constraints considered in this work). To allow for parallel use of Python / SymPy from within Julia, PySR was run in distributed mode, necessitating more overhead than threaded mode.

\end{document}